\documentclass[a4paper,11pt]{article}
\usepackage{stmaryrd}
\usepackage{graphicx}
\usepackage{amsfonts}
\usepackage{amsmath,amsthm}
\usepackage{amssymb}
\usepackage{mathrsfs}
\usepackage{indentfirst}
\usepackage{titlesec}
\usepackage{caption2}
\usepackage{color}
\usepackage[normalem]{ulem}
\usepackage{algorithm}
\usepackage{algorithmic}

\usepackage[english]{babel}
\usepackage{cite}
\usepackage{bbm}

\newtheorem{theorem}{Theorem}[section]
\newtheorem{definition}{Definition}[section]

\newtheorem{remark}[theorem]{Remark}
\newtheorem{lemma}[theorem]{Lemma}
\newtheorem{corollary}[theorem]{Corollary}
\newtheorem{example}{Example}[section]

\newtheorem{assumption}{Assumption}

\providecommand{\1}{\mathbf 1}

\begin{document}

\title{Quasi--potential as an implicit regularizer for the loss function in the stochastic gradient descent.}

\author{Wenqing Hu
\thanks{Department of Mathematics and Statistics, Missouri University of Science and Technology
(formerly University of Missouri, Rolla). Email: \texttt{huwen@mst.edu}} \ ,
Zhanxing Zhu
\thanks{Peking University and Beijing Institute of Big Data Research, Beijing, China.
Email: \texttt{zhanxing.zhu@pku.edu.cn}} \ ,
Haoyi Xiong
\thanks{Big Data Lab, Baidu Inc. and National Engineering Laboratory of Deep Learning Application and Technology.
Email: \texttt{xhyccc@gmail.com}}\ ,
and Jun Huan
\thanks{Big Data Lab, Baidu Inc. and National Engineering Laboratory of Deep Learning Application and Technology.
Email: \texttt{huanjun@baidu.com}}}

\date{\today}

\maketitle

\begin{abstract}
We interpret the variational inference of the Stochastic Gradient Descent (SGD) as minimizing a new potential function named
the \textit{quasi--potential}. We analytically construct the quasi--potential function in the case when the loss function is convex and admits only one
global minimum point. We show in this case that the quasi--potential function is related to the noise covariance structure
of SGD via a partial differential equation of Hamilton--Jacobi type. This relation
helps us to show that anisotropic noise leads to faster escape than isotropic noise.
We then consider the dynamics of SGD in the case when the loss function is non--convex and admits several different local minima.
In this case, we demonstrate an example that shows how the noise covariance structure plays a role
in ``implicit regularization", a phenomenon in which SGD favors some particular local minimum points.
This is done through the relation between the noise covariance structure and the quasi--potential function. Our analysis is based on
Large Deviations Theory (LDT), and they are validated by numerical experiments.
\end{abstract}

\section{Introduction.}

The statistical performance of the Stochastic Gradient Descent (SGD) and its variants
has led to effective training of large--scale statistical learning models. It is widely believed that
 SGD is an ``implicit regularizer" which helps itself to search for a local minimum point that is easy to generalize
 (see \cite{[Zhang2016RethinkingDeepLearning]},
 \cite{[NeyshaburEtAl2017ImplicitRegularization]}, \cite{[ShwarzTishbyDNN2017]}).
This belief has been stemming from its remarkable empirical performance.
To justify this belief from a solid theoretical perspective, optimization behaviors of SGD and their
impacts on generalization has been studied from attempts of
characterizing how SGD escapes from stationary points, including
saddle points and local minima (see \cite{[JinChiEtAlTensorDecomposition]}, \cite{[JinChiEtALEscapeSaddlePoint]}).
Another stream of research focuses on interpreting SGD as performing variational inference, in which it is found that
SGD minimizes the Kullback--Leibler divergence between the stationary distribution to its posterior
(see \cite{[BleiSGDVariationalInference]}, \cite{[MandtEtAlSGDVariationalICML2016]}, \cite{[Duvenaud2016EarlyStoppingAISTATS]}).
Along this direction, it is realized that SGD performs a variational inference using a new potential function that it implicitly
constructs given an architecture and a dataset.

In the above mentioned researches, ``implicit regularization" was explained to be related to the
``noise tied to architecture" (see \cite[Section 6]{[Chaudhari-SoattoSGDVariational]})
arising in, e.g. dropout or small mini--batches used in SGD. For example in variational inference approach
the new potential function, denoted as $\Phi$, can be shown to be equal up to a scalar multiple
to the original loss function in the case when the noise in SGD is isotropic
(see \cite[Lemma 6]{[Chaudhari-SoattoSGDVariational]}).
However in real case, due to the special
architecture of deep networks where gradient noise is highly non--isotropic
(anisotropic, see \cite[Section 4.1]{[Chaudhari-SoattoSGDVariational]}, \cite{[HofmannEtAlEscapeSaddleSGD]}, \cite{[ZhanxingSGDAnisotropic]}),
it is believed that the potential $\Phi$ possesses properties
that lead to both generalization and acceleration (see \cite{[Duvenaud2016EarlyStoppingAISTATS]}).
From the perspective of escaping from stationary points, it is also found in \cite{[ZhanxingSGDAnisotropic]}
that anisotropic noise leads to faster escape from sharp local minima, which validates previous results from
both empirical and theoretical analysis
(see \cite{[SGDEscapeLocalMinima]}, \cite{[HofmannEtAlEscapeSaddleSGD]}, \cite{[KeskarEtAlLargeBatchTraining]},
\cite{[HittingTimeSGLD]}).

In this work, we provide a unified approach to these problems.
Instead of the potential function $\Phi$, we will
construct another potential function -- the \textit{(global) quasi-potential} $\phi^{QP}$ that characterizes the long--time behavior of SGD with small learning rate.
We demonstrate that, as the learning
rate tends to zero, SGD finally enters the global minimum of the (global) quasi--potential function $\phi^{QP}$.
It is interesting to observe that in the case when the loss landscape possesses only
one global minimum and under isotropic noise, the (global) quasi--potential
that we constructed  becomes a local quasi--potential and it agrees
with the original loss function up to a multiplicative constant.
However, with the presence of
multiple potential wells,
SGD with anisotropic noise minimizes a (global) quasi--potential
that has a different landscape from the original loss function.

The construction of the (global) quasi--potential function is done by first calculating the local quasi--potential function,
which is valid within one basin of attraction around a specific local minimum point. Analytically,
this local quasi--potential function is the solution of a variational problem in which the Lagrangian has explicit dependence
on the covariance structure of the noise (e.g. provided by the architecture of the neural network) and the loss landscape.
By classical calculus of variations, such a function can be calculated from a partial differential equation of
 Hamilton--Jacobi type with prescribed boundary value condition.
From this perspective one can explicitly see the dependence of the quasi--potential on the covariance structure of the noise.
This suggests that the quasi--potential can capture the effects of implicit regularization brought in by
the ``noise tied to architecture" from algorithms like SGD. In particular,
highly anisotropic noise in SGD also induces a (global) quasi--potential function
that will be different from the original loss function, and this is more relevant to the training of real deep neural networks.

Our results are in the case when the learning rate asymptotically tends to zero. This is different when
compared to the previous work like
\cite{[Chaudhari-SoattoSGDVariational]} that considers the case of moderate learning rate.
We demonstrate that even in this case, implicit regularization still exists and manifests itself through the (global)
quasi--potential function $\phi^{QP}$.
This is mainly due to the exponentially long escape time from local minima induced by small noise in SGD.
Our analysis relies on Large Deviations Theory (LDT) (see \cite{[FWbook1998]}, \cite{[FWbook2012]}, \cite{[Dembo-Zeitouni]}),
in which exponentially small transition probabilities are estimated
through a path--integral along trajectories exiting from local minima.
These exponents depend on the learning rate as well as the noise covariance structure.
Via the Markov property of the SGD process, such exponentially small exit probabilities can be turned into estimates of the
exponentially long escape time from the basin of attractors around local minima.
As the exponential complexity of SGD based algorithms has also been
noted in \cite{[JinEtAlPolulation-EmpiricalNIPS2018]} via information--theoretic guarantees, our work demonstrates that
the exponentially long escape time from local minima induced by small noise in SGD should be a key factor that leads to implicit regularization.

Although our constructed (global) quasi--potential function $\phi^{QP}$ is different compared to the potential function $\Phi$ constructed
via variational inference (see \cite{[Chaudhari-SoattoSGDVariational]}), they are inter--connected via the steady--state distribution
of the Fokker--Plank equation corresponding to the stochastic dynamics of SGD. In fact the potential function $\Phi$ constructed
in variational inference comes directly from the exponent in the exponential form of the
density of the stationary--state distribution. This has been the case when the learning rate (step--size of the recursive scheme)
is moderate. In the case when the learning rate is small, the normalizing constant
(or in the language of statistical physics, the partition function) also scales with the learning rate and will affect the exponent from which
we calculate the previous potential function $\Phi$. At the limit when the learning rate tends to zero,
this leads to our new potential function, which is the (global)
quasi--potential $\phi^{QP}$.

Another interesting point here is that from our probabilistic considerations,
we can understand not only the ``final" behavior of the SGD algorithm
in the long--time, that is which minimum is going to be finally achieved,
but also the dynamics of the SGD algorithm jumping between different local minima at an intermediate time scale.
Such ``metastable" behavior can be characterized by a Markov chain with exponentially small transition
probabilities between local minima of the quasi--potential $\phi^{\text{QP}}$. This Markov chain is a reflection of
the procedure during which SGD selects those specific local minimum points that it favors
(such as those with good generalization properties). From here we can characterize the mechanism of implicit regularization
by this Markov chain associated with the quasi--potential. We will illustrate this point via an example
(Example \ref{Example:TwoWellPotentialLossFunctionGlobalQuasipotential}) in Section \ref{Sec:GlobalQuasiPotential}.

The paper is organized as follows. We will review in Section \ref{Sec:Background} basic information of continuous SGD
and variational inference. We will then demonstrate in Section \ref{Sec:LocalQuasiPotential}
the construction of the local quasi--potential function, in particular how this quantity is related
to the noise covariance matrix (sometimes also referred as the diffusion matrix)
via a partial differential equation of Hamilton--Jacobi type. We accompany this section with an example showing that
when the diffusion matrix is anisotropic, the escape from a local minimum can be faster than the case with isotropic noise.
We further demonstrate the construction of the global quasi--potential
via an example in Section \ref{Sec:GlobalQuasiPotential},
where the loss function is given by a two--well potential in dimension $2$, and the diffusion matrix is anisotropic.
We illustrate the construction of the (global) quasi--potential and we show that it has a different landscape from the original loss function.
We also demonstrate via this example the metastable dynamics of SGD, that is a Markov chain between different local minimum points.
This Markov chain leads to SGD's final trapping into the global minimum of the quasi--potential, that may be different from the global minimum
of the original loss function.
In Section \ref{Sec:Numerics} we provide numerical results for both our examples in Sections \ref{Sec:LocalQuasiPotential} and
\ref{Sec:GlobalQuasiPotential}. Finally, we conclude this work and we propose future directions in Section \ref{Sec:Conclusion}.

\

\textbf{Main contributions of this paper.}
In this work, a new potential function named the \textit{quasi--potential} was introduced and the
variational inference of SGD was interpreted as minimizing the quasi--potential function.
By making use of LDT and classical calculus of variations,
a relation between the quasi--potential and the noise covariance structure (the diffusion matrix in SGD)
is revealed through a partial differential equation of Hamilton--Jacobi type.
This relation helps to show that anisotropic noise leads to faster escape than isotropic noise.
Furthermore, the mechanism of ``implicit regularization" was explained through a Markov chain
between local minimum points of the quasi--potential.
This Markov chain is induced by the noise in SGD and it is associated with
the noise covariance structure via the relation between the diffusion matrix and the quasi--potential.
This work proposes a quantitative way to understand the phenomenon of ``implicit regularization"
by proposing the quasi--potential and relate it to the noise covariance structure via partial differential equation
and stochastic dynamics.

\

\textbf{Comparison with previous works.}
There has been a large literature dedicated to the empirical fact that SGD favors local minimum points that have
good empirical generalization properties (see for example \cite{[Zhang2016RethinkingDeepLearning]},
 \cite{[NeyshaburEtAl2017ImplicitRegularization]}, \cite{[ShwarzTishbyDNN2017]}). On the theoretical side, attempts have
 been made to explain this ``implicit regularization" associated with the SGD process
 through its stochastic dynamics, such as escape from stationary points
 (see \cite{[JinChiEtAlTensorDecomposition]}, \cite{[JinChiEtALEscapeSaddlePoint]}, \cite{[SGDEscapeLocalMinima]}, \cite{[HofmannEtAlEscapeSaddleSGD]}, \cite{[KeskarEtAlLargeBatchTraining]}, \cite{[HittingTimeSGLD]}).
Variational inference has been discussed in such works as
\cite{[BleiSGDVariationalInference]}, \cite{[MandtEtAlSGDVariationalICML2016]}, \cite{[Duvenaud2016EarlyStoppingAISTATS]}.
There has been attempts trying to relate the covariance structure of the SGD noise
and its generalization properties (see \cite{[Chaudhari-SoattoSGDVariational]}, \cite{[HofmannEtAlEscapeSaddleSGD]}),
in particular how anisotropic noise leads to fast escape from saddle points and local minimum points (see \cite{[HofmannEtAlEscapeSaddleSGD]}, \cite{[ZhanxingSGDAnisotropic]}). Compared to these previous works, our work proposed a unified way to
enhance the understanding of the connection between SGD's noise covariance structure and its selection of specifically favored local minimum points.
The novelty here is that the quasi--potential function that we introduced can be quantitatively related to SGD's
noise covariance structure via a partial differential equation of Hamilton--Jacobi type. This provides us with a general analytic tool
to compare the effects of isotropic v.s. anisotropic noise. The derivation of our analysis based on LDT also proposes a
further understanding of the mechanism behind implicit regularization. Indeed, from LDT
we understand that SGD selects its favorite local minimum points through performing a Markov chain
between different local minima, and the behavior of this Markov chain is related to SGD's noise covariance structure.

\section{Background on continuous--time SGD and the stationary distribution, statement of the problem.}
\label{Sec:Background}

\subsection{Continuous--time SGD.}
\label{Sec:Background:SubSec:ContinuousTimeSGD}

Stochastic gradient descent (see \cite{[BottouSGD2010]}, \cite{[Bach-MoulinesSGD2013NIPS]}, \cite{[Flammarion-BachCOLT2015]})
with a constant learning rate
is a stochastic analogue of the gradient descent algorithm, aiming at finding
the local or global minimizers of the function expectation parameterized by some random variable.
Schematically, the algorithm can be interpreted as targeting at finding a local minimum point of
the expectation of function
\begin{equation}\label{Eq:OptimizationObjective}
f(x)=\mathbf{E} f(x;\zeta) \ ,
\end{equation}
where the index random variable $\zeta$ follows some
prescribed distribution $\mathcal{D}$, and the weight vector $x\in \mathbb{R}^d$.
The stochastic gradient descent updates via the iteration

\begin{equation}\label{Eq:StochasticGradientDescent}
x_{k+1}=x_k-\eta \nabla f(x_k; \zeta_k) \ ,
\end{equation}
where $\eta>0$ is a fixed step--size which is also the learning rate, and $\zeta_k$ are i.i.d. random variables
that have the same distribution as $\zeta$. In particular, in the case of training a deep neural network,
the random variable $\zeta$ samples size $m$ ($m\leq n$) mini--batches $\mathcal{B}$ uniformly from an index set $\{1,2,...,n\}$:
$\mathcal{B}\subset \{1,2,...,n\}$ and $|\mathcal{B}|=m$. In this case, given loss functions $f_1(x), ..., f_n(x)$ on
training data, we have $f(x; \zeta)=\dfrac{1}{m}\sum\limits_{i\in \mathcal{B}}f_i(x)$ and $f(x)=\dfrac{1}{n}\sum\limits_{i=1}^n f_i(x)$.
Set
\begin{equation}\label{Eq:Epsilon}
\varepsilon=\dfrac{\eta}{m} \ .
\end{equation}
Based on the iteration \eqref{Eq:StochasticGradientDescent}, we introduce a stochastic differential equation (SDE)
for the discrete--time SGD updates

\begin{lemma}\label{Lemma:SDE-SGD}
The continuous--time limit of SGD is given by
\begin{equation}\label{Lemma:SDE-SGD:Eq:SDE}
dx(t)=-\nabla f(x(t))dt+\sqrt{\varepsilon} \Sigma(x(t))dW(t) \ ,
\end{equation}
where $W(t)$ is a standard Brownian motion in $\mathbb{R}^d$ and the matrix
$\Sigma(x)$ satisfies $\Sigma(x)\Sigma^T(x)=D(x)$, where the diffusion matrix $D(x)$ is the
nonnegative--definite matrix
\begin{equation}\label{Lemma:SDE-SGD:Eq:DiffusionMatrix}
D(x)=\mathbf{E}(\nabla f(x; \zeta)-f(x))(\nabla f(x; \zeta)-f(x))^T \ .
\end{equation}
\end{lemma}

We refer to \cite{[WeinanEtAlSDE2017]},  \cite{[WeinanEtAlSDE2018]}, \cite{[JunchiEtAlSDE2017]}, \cite{[Hu-LiSGD]}, \cite{[Hu-Li-SuNesterov]}, \cite{[Hu-Li-SCGD]} for the proof of the convergence of discrete SGD \eqref{Eq:StochasticGradientDescent} to \eqref{Lemma:SDE-SGD:Eq:SDE}. The diffusion matrix $D(x)$ depends on the weight vector $x$, the architecture of the learning model
(such as a neural network) and the loss function defined by $f(x)$ as well as the data set. When $D(x)=cI_d$
is a scalar multiple of the identity, independent of $x$, we call $D(x)$ an \textit{isotropic} diffusion matrix; otherwise, we call
$D(x)$ \textit{non--isotropic} (\textit{anisotropic}). In real case when the architecture is given by
a deep neural network, the diffusion matrix $D(x)$ is usually anisotropic with a large condition number
(see \cite{[Chaudhari-SoattoSGDVariational]}, \cite{[ZhanxingSGDAnisotropic]}).

For mathematical reasons we will make the following simple assumptions regarding the loss function $f(x)$ and the diffusion matrix $D(x)$.

\begin{assumption}\label{Assumption:SGDLossfDiffusionD}
We assume that the loss function $f(x)$ admits a gradient $\nabla f(x)$ which is $L$--Lipschitz
\begin{equation}\label{Assumption:SGDLossfDiffusionD:Eq:LossfLLipschitz}
|\nabla f(x)-\nabla f(y)|\leq L|x-y| \text{ for all } x, y\in \mathbb{R}^d \text{ and some } L>0 \ .
\end{equation}
We assume that $\Sigma(x)$ is piecewise Lipschitz in $x$ and the diffusion matrix $D(x)$ is invertible for all $x\in \mathbb{R}^d$ such that
\begin{equation}\label{Assumption:SGDLossfDiffusionD:Eq:DiffusionDBoundedTrace}
\text{Tr} D(x) \leq M \text{ for all } x\in \mathbb{R}^d \text{ and some } M>0 \ .
\end{equation}
\end{assumption}
Here and below $\text{Tr}$ is the trace operator applied to a square matrix.

\begin{remark}\label{Remark:AnisotropicDiffusion}
Although we assume here that the diffusion matrix $D(x)$ is invertible for all choices of $x\in \mathbb{R}^d$,
this does not exclude the case that it is anisotropic
or in other words it admits a large condition number.
\end{remark}

\subsection{Steady--State Distribution and Variational Inference.}
\label{Sec:Background:SubSec:VariationalInference}

The steady--state distribution of the weights $x(t)$, given by a density function
$\rho(x,t)dx\propto \mathbf{P}(x(t)\in dx)$, evolves according to the Fokker--Planck equation
(see \cite[Chapter 8]{[OksendalSDE]})

\begin{equation}\label{Eq:FokkerPlank}
\dfrac{\partial \rho}{\partial t}=\nabla\cdot\left(\nabla f(x) \rho\right)+\dfrac{\varepsilon}{2}\text{Tr}(D(x) \nabla^2\rho) \ .
\end{equation}

We make an assumption
on the uniqueness of stationary--state distribution.

\begin{assumption}\label{Assumption:UniquenessStationaryMeasure}
We assume that the steady--state distribution of the Fokker--Plank equation exists and is unique. We denote the density of the
stationary state distribution to be $\rho^{\text{SS}}(x)$ and it satisfies
\begin{equation}\label{Assumption:UniquenessStationaryMeasure:Eq:DensitySS}
0=\dfrac{\partial \rho^{\text{SS}}}{\partial t}=\nabla\cdot\left(\nabla f(x) \rho^{\text{SS}}\right)
+\dfrac{\varepsilon}{2}\text{Tr}(D(x) \nabla^2\rho^{\text{SS}}) \ .
\end{equation}
\end{assumption}

In the variational inference interpretation of SGD (see \cite{[BleiSGDVariationalInference]}, \cite{[MandtEtAlSGDVariationalICML2016]}, \cite{[Duvenaud2016EarlyStoppingAISTATS]}), an implicitly defined potential $\Phi(x)$ was introduced (see \cite{[Chaudhari-SoattoSGDVariational]}) using the
steady--state distribution $\rho^{\text{SS}}$ as
\begin{equation}\label{Eq:VariationalInferencePotentialPhi}
\Phi(x)=-\dfrac{\varepsilon}{2}\ln \rho^{\text{SS}}(x)
\end{equation}
up to a multiplicative constant. In this way, the stationary density $\rho^{\text{SS}}$ can be expressed in terms of the
potential function $\Phi$ using a normalizing constant $Z(\varepsilon)$, which is the partition function in statistical physics, as

\begin{equation}\label{Eq:StationaryDensityInTermsOfPhi}
\rho^{\text{SS}}(x)=\dfrac{1}{Z(\varepsilon)}\exp\left(-\frac{2}{\varepsilon}\Phi(x)\right) \ .
\end{equation}

Under the Assumption 1, that is the existence and uniqueness of a stationary (invariant) density $\rho^{\text{SS}}$, it is
guaranteed that the as time $t\rightarrow \infty$, the SGD distribution density function $\rho(x,t)$ converges to $\rho^{\text{SS}}(x)$
in the sense of KL divergence. This is the variational inference interpretation of SGD. We have the following Theorem proved in
\cite[Theorem 5]{[Chaudhari-SoattoSGDVariational]}.

\begin{theorem}\label{Thm:SGDVariationalInference}
The functional $F(\rho)=\dfrac{\varepsilon}{2}\text{KL}(\rho||\rho^{\text{SS}})$ decreases monotonically along the trajectories
of the Fokker--Plank equation \eqref{Eq:FokkerPlank} as $t\rightarrow \infty$
and converges to its minimum, which is zero, at steady--state. Here
$$\text{KL}(\rho||\rho^{\text{SS}})=\displaystyle{\int_{\mathbb{R}^d}}\rho(x,t)\ln\left(\dfrac{\rho(x,t)}{\rho^{\text{SS}}(x)}\right)dx$$
is the KL divergence between $\rho$ and $\rho^{\text{SS}}$. Further, an energy--entropy split for the functional $F$ is given by
$$F(\rho)=\mathbf{E}_{x\in \rho}[\Phi(x)]-\dfrac{\varepsilon}{2}H(\rho)+\text{constant} \ ,$$
where $H(\rho)=-\displaystyle{\int_{\mathbb{R}^d}\rho\ln\rho}$ is the entropy of the distribution $\rho$.
\end{theorem}

In particular, the above Theorem implies the following Corollary.

\begin{corollary}\label{Cor:SGD-DynamicsAsymptoticTimeInvMeasure}
We have $\rho(x,t)\rightarrow \rho^{\text{SS}}(x)$ as $t\rightarrow\infty$ for every $x\in \mathbb{R}^d$.
\end{corollary}

Let us fix $\eta$ and $m$ and thus we consider the case
when $\varepsilon=\dfrac{\eta}{m}>0$ is a fixed parameter. In this
case, based on the above Theorem, \cite{[Chaudhari-SoattoSGDVariational]} shows
that the steady--state of SGD in \eqref{Eq:StationaryDensityInTermsOfPhi}
is such that it places most of its probability mass
in regions of the parameter space with small values of $\Phi$.
In this sense, the potential function $\Phi=\Phi(x)$ is
understood as a new objective function, instead of the original
function $f(x)$ that SGD tries to minimize in \eqref{Eq:OptimizationObjective}
(see more discussions in \cite{[Chaudhari-SoattoSGDVariational]}). In particular, the function $\Phi(x)$
captures information from both the original loss function $f(x)$ and the diffusion matrix $D(x)$.
This has been discussed as a manifestation of implicit regularization via the SGD trajectory (see \cite{[Chaudhari-SoattoSGDVariational]}).
In particular, if $D(x)=cI_d$ and without boundary condition, then one can show that $\Phi(x)=\widetilde{c}f(x)$ (see \cite[Lemma 6]{[Chaudhari-SoattoSGDVariational]}),
so that isotropic diffusion will not bring in new effects via implicit regularization.

\subsection{The asymptotic as $\varepsilon\rightarrow 0$ and the quasi--potential.}
\label{Sec:Background:SubSec:QuasiPotential}

The above considerations are in the case when $\varepsilon>0$ is fixed, rather than the case when $\varepsilon\rightarrow 0$.
As $\varepsilon\rightarrow 0$ is small, the normalizing factor $Z(\varepsilon)$ in \eqref{Eq:StationaryDensityInTermsOfPhi}
also scales with $\varepsilon$ and results in the fact that the asymptotic of the stationary distribution $\rho^{\text{SS}}$
does not depend only on the potential function $\Phi$.

\begin{definition}\label{Def:LogarithmicEquivalence}
\emph{(logarithmic equivalence)}
Two families of quantities $A(\varepsilon)>0$ and $B(\varepsilon)>0$ that depend on $\varepsilon>0$ are said to be logarithmically equivalent,
and denoted by
$$A(\varepsilon)\asymp B(\varepsilon) \ ,$$
if and only if
\begin{equation}\label{Def:LogarithmicEquivalence:Eq:Limit}
\lim\limits_{\varepsilon\rightarrow 0}\varepsilon[\ln A(\varepsilon)-\ln B(\varepsilon)]=0 \ .
\end{equation}
Or in other words, for any $\gamma>0$ there exists some $\varepsilon_0>0$ such that we have
\begin{equation}\label{Def:LogarithmicEquivalence:Eq:EpsDelta}
\exp\left(-\dfrac{\gamma}{\varepsilon}\right)B(\varepsilon)
\leq
A(\varepsilon)
\leq
\exp\left(\dfrac{\gamma}{\varepsilon}\right)B(\varepsilon) \ ,
\end{equation}
for any $0<\varepsilon<\varepsilon_0$.
\end{definition}

Our goal in this paper is to argue that as $\varepsilon$
is small and is close to $0$ we have
\begin{equation}\label{Eq:DensityInvariantMeasureExponentialQuasiPotential}
\rho^{\text{SS}}(x) \asymp \exp\left(-\dfrac{1}{\varepsilon}\phi^{\text{QP}}(x)\right) \ .
\end{equation}
That is, for any $\gamma>0$ we can pick an $\varepsilon_0>0$
small enough such that
$$\exp\left(-\dfrac{1}{\varepsilon}\left(\phi^{\text{QP}}(x)+\gamma\right)\right)\leq
\rho^{\text{SS}}(x)
\leq \exp\left(-\dfrac{1}{\varepsilon}\left(\phi^{\text{QP}}(x)-\gamma\right)\right)$$
for all $0<\varepsilon< \varepsilon_0$. Or in other words
$$-\lim\limits_{\varepsilon\rightarrow 0}\varepsilon\ln \rho^{\text{SS}}(x)=\phi^{\text{QP}}(x) \ .$$
The function $\phi^{\text{QP}}(x)$ is called the \textit{global quasi--potential function}
(or sometimes abbreviated as the quasi--potential function depending on the context)
and can be constructed from the original loss function $f(x)$ and the diffusion matrix $D(x)$.
Thus it depends on the weight vector $x$, the architecture of the learning model
(such as a neural network) and the loss function defined by $f(x)$ as well as the data set.

Asymptotic identity \eqref{Eq:DensityInvariantMeasureExponentialQuasiPotential} does not involve any normalizing constant
$Z(\varepsilon)$ as in \eqref{Eq:StationaryDensityInTermsOfPhi}, and indeed the global quasi--potential function
$\phi^{\text{QP}}(x)$ has a global minimum $x^*\in \mathbb{R}^d$ such that $\phi^{\text{QP}}(x^*)=0$. This combined with the ansatz
\eqref{Eq:DensityInvariantMeasureExponentialQuasiPotential} indicates that as $\varepsilon$ is small, the stationary density
$\rho^{\text{SS}}(x)$ will be concentrated on a certain global minimum point $x^*$ of the quasi--potential $\phi^{\text{QP}}(x)$.
By Corollary \ref{Cor:SGD-DynamicsAsymptoticTimeInvMeasure}, we see that this global minimum point $x^*$
can be understood as the long--time behavior of SGD dynamics as first $t\rightarrow\infty$ and then $\varepsilon\rightarrow 0$.
This indicates that in the asymptotic regime when $\varepsilon\rightarrow 0$,
SGD minimizes the quasi--potential $\phi^{\text{QP}}(x)$ rather than the original function $f(x)$.

Comparing \eqref{Eq:StationaryDensityInTermsOfPhi} and \eqref{Eq:DensityInvariantMeasureExponentialQuasiPotential},
we see that

\begin{equation}\label{Eq:Relation:QuasiPotential-VariationalInference}
\left|\phi^{\text{QP}}(x)-2\Phi(x)-\varepsilon\ln Z(\varepsilon)\right|\rightarrow 0
\end{equation}
as $\varepsilon\rightarrow 0$. Thus we see that the two potential functions $\Phi(x)$ and $\phi^{\text{QP}}(x)$ differ by a term $\varepsilon \ln Z(\varepsilon)$ that
involves the normalizing factor (partition function) $Z(\varepsilon)$. Moreover, for fixed $\varepsilon>0$, the potential $\Phi(x)$ may depend on $\varepsilon$,
and thus we can think of $\phi^{\text{QP}}(x)$ as the limit $\phi^{\text{QP}}(x)=\lim\limits_{\varepsilon\rightarrow 0} \left(2\Phi(x)+\varepsilon\ln Z(\varepsilon)\right)$.

In the next two sections we will demonstrate how $\phi^{\text{QP}}(x)$ is calculated via the diffusion matrix $D(x)$ and the loss landscape
$f(x)$.

\section{Local quasi--potential: the case of convex loss function.}
\label{Sec:LocalQuasiPotential}

Let us assume in this Section that the original loss function $f(x)$ is convex and
admits only one minimum point $O$, which is also its global minimum point. Let $O$ be the origin.
In this Section we will introduce the local quasi--potential function and we will connect it to the
SGD noise covariance structure via a partial differential equation of Hamilton--Jacobi type.
The analysis is based on interpreting the LDT as a path integral theory in the trajectory space.

\subsection{SGD as a small random perturbation of Gradient Descent (GD).}
\label{Sec:LocalQuasiPotential:SubSec:SGD-Perturbation-GD}

For small $\varepsilon>0$ the SGD process $x(t)$ in \eqref{Lemma:SDE-SGD:Eq:SDE} has trajectories
that are close to the Gradient Descent (GD) flow
characterized by the deterministic equation
\begin{equation}\label{Eq:GradientSystem}
\dfrac{dx^{\text{GD}}(t)}{dt}=-\nabla f(x^{\text{GD}}(t)) \ , \ x(0)=x_0 \ .
\end{equation}
In fact, it can be easily justified (see Appendix \ref{Appendix:PerturbationGD-SGD}) that we have the following

\begin{lemma}\label{Lemma:SGD-GD-Closeness}
Under Assumption \ref{Assumption:SGDLossfDiffusionD} we have, for any $T>0$,
\begin{equation}\label{Lemma:SGD-GD-Closeness:Eq:SqrtEpsClosenessSGD-GD}
\max\limits_{0\leq t\leq T}\mathbf{E}|x(t)-x^{\text{GD}}(t)|^2\leq C\varepsilon
\end{equation}
for some constant $C=C(T, L, M)>0$.
\end{lemma}

When \eqref{Lemma:SGD-GD-Closeness:Eq:SqrtEpsClosenessSGD-GD} holds, we will simply say that
$x(t)$ and $x^{\text{GD}}(t)$ are $\mathcal{O}(\sqrt{\varepsilon})$--close on $0\leq t \leq T$.
Thus in finite time the SGD process $x(t)$ will be attracted to a neighborhood of the origin $O$.
Since $O$ is the only minimum point of the convex loss function $f(x)$, every point in $\mathbb{R}^d$
is attracted by the gradient flow \eqref{Eq:GradientSystem} to $O$.
Let us take any open set $U$ containing the origin $O$.
Due to the attraction property of the deterministic gradient descent flow $x^{\text{GD}}(t)$ and the $\mathcal{O}(\sqrt{\varepsilon})$--closeness
of the SGD path $x(t)$ from $x^{\text{GD}}(t)$, the SGD solution process $x(t)$ will then spend a long time staying in this
neighborhood $U$ of the origin $O$,
before it escapes from $U$ and hits somewhere on $\partial U$. Such an
escape is due to small random term $\sqrt{\varepsilon}\Sigma(x(t))dW(t)$ in \eqref{Lemma:SDE-SGD:Eq:SDE}, that leads to fluctuations
of the SGD process $x(t)$ deviating from the deterministic trajectory of the gradient descent flow $x^{\text{GD}}(t)$.
In terms of optimization, SGD in this case finds the minimum point $O$ of the convex function $f(x)$ just as GD does. However,
at the presence of multiple local minimum points,
the escape from $O$ due to small randomness in SGD solution process \eqref{Lemma:SDE-SGD:Eq:SDE}
 is a key feature that leads to its regularization properties, such as the selection
of flat minimizers against sharp minimizers (see \cite{[KeskarEtAlLargeBatchTraining]}).
The understanding of escape properties from the basin of attractors
due to small random perturbations can also be performed in the case of just one minimum point $O$.
In this case, we can take an open neighborhood $U$ of $O$ and consider escape behavior of the SGD process $x(t)$
from the set $U$ to its boundary $\partial U$. This will be done via LDT in the next subsection.

\subsection{Large Deviations Theory (LDT) interpreted as a path integral in the trajectory space.}
\label{Sec:LocalQuasiPotential:SubSec:LDT-PathIntegral}

To quantitatively characterize such escape properties, we propose to use \textit{Large Deviations Theory} (LDT)
(see \cite{[FWbook1998]}, \cite{[FWbook2012]}, \cite{[Dembo-Zeitouni]}). Roughly, this theory gives the probability weights in the path space
of the solution $x(t)$ of \eqref{Lemma:SDE-SGD:Eq:SDE}. That is to say, for a given regular connecting path $\psi(t)$,
$\psi(0)=x(0)=x_0$, $\psi(T)=x(T)$, and some $\delta>0$ small enough, for any $\varepsilon>0$
small enough, we have

\begin{equation}\label{Eq:LDPEstimate}
\mathbf{P}\left(\sup\limits_{0\leq t\leq T}|x(t)-\psi(t)|\leq \delta\right)\asymp \exp\left(-\dfrac{1}{\varepsilon}S_{0T}(\psi)\right) \ ,
\end{equation}
where $\asymp$ denotes logarithmic equivalence. The asymptotic \eqref{Eq:LDPEstimate} can be understood
as providing a density function for the process $x(t)$ in the path space:

\begin{equation}\label{Eq:LDPPathSpaceDensity}
\mathbf{P}(\psi|\psi(0)=x_0, \psi(T)=x)\asymp \exp\left(-\dfrac{1}{\varepsilon}S_{0T}(\psi)\right) \ .
\end{equation}

The precise statement that leads to the asymptotic
\eqref{Eq:LDPEstimate} will be illustrated in Appendix \ref{Appendix:ElementsLDP}.
In LDT, for the asymptotic \eqref{Eq:LDPEstimate},
the functional $S_{0T}(\psi)$ is called the \textit{rate functional}. However, this rate functional can be
interpreted as the \textit{action functional} that gives the solution to the Fokker--Plank equation
\eqref{Eq:FokkerPlank}. In fact, following Feynman's path integral approach to quantum mechanics
(see \cite{[FeynmannPathIntegral]}, \cite{[Li-ZhouAIntegral]}), formally by integrating the individual
paths $\psi(t)$ according to their weights given in
\eqref{Eq:LDPPathSpaceDensity}, we have that

\begin{equation}\label{Eq:FeynmanPathIntegral}
\begin{array}{ll}
\rho(x,T|x_0,0)&=\displaystyle{\int \mathcal{D} \psi \mathbf{P}(\psi|\psi(0)=x_0, \psi(T)=x)}
\\
&\asymp\displaystyle{\int \mathcal{D} \psi \exp\left(-\dfrac{1}{\varepsilon}S_{0T}(\psi)\right)} \ ,
\end{array}
\end{equation}
where the integral is a formal integration on the path space with ``path differential" $\mathcal{D}\psi$, and
$\rho(x,T|x_0,0)$ is the solution to the Fokker--Plank equation \eqref{Eq:FokkerPlank} (it is a partial differential equation)
with initial density $\rho(x,0)=\delta(x-x_0)$. In Feynman's theory, the term ``action" corresponds to the exponent
$S_{0T}(\psi)$ in the above integration. In this way from \eqref{Eq:FeynmanPathIntegral} we get

\begin{equation}\label{Eq:LaplaceMethodFeynmanPathIntegral}
\begin{array}{ll}
&-\lim\limits_{\varepsilon\rightarrow 0}\varepsilon \ln\rho(x,T|x_0,0)
\\
=& -\lim\limits_{\varepsilon\rightarrow 0}\varepsilon \displaystyle{\ln \int \mathcal{D} \psi \exp\left(-\dfrac{1}{\varepsilon}S_{0T}(\psi)\right)}
\\
\stackrel{(*)}{=}& \inf\limits_{\psi(0)=x_0, \psi(T)=x}S_{0T}(\psi) \ .
\end{array}
\end{equation}
Here the step $(*)$ follows the classical Laplace's method in the approximate integral theory (see \cite{[Laplace1774]}).
From Corollary \ref{Cor:SGD-DynamicsAsymptoticTimeInvMeasure} we have

$$-\lim\limits_{\varepsilon\rightarrow 0}\varepsilon\ln\rho^{\text{SS}}(x)
=-\lim\limits_{\varepsilon\rightarrow 0}\lim\limits_{T\rightarrow\infty}\varepsilon\ln \rho(x,T|x_0,0) \ .$$

Assuming in the above that the limit $\varepsilon\rightarrow 0$ and $T\rightarrow \infty$ are interchangeable, then we have
by \eqref{Eq:LaplaceMethodFeynmanPathIntegral} that

\begin{equation}\label{Eq:StationaryMeasureExponentialForm}
\begin{array}{ll}
-\lim\limits_{\varepsilon\rightarrow 0}\varepsilon\ln\rho^{\text{SS}}(x)
&= -\lim\limits_{T\rightarrow\infty}\lim\limits_{\varepsilon\rightarrow 0}\varepsilon\ln \rho(x,T|x_0,0)
\\
&= \lim\limits_{T\rightarrow\infty}\left(\inf\limits_{\psi(0)=x_0, \psi(T)=x}S_{0T}(\psi)\right)
\\
&= \inf\limits_{T>0}\inf\limits_{\psi(0)=x_0, \psi(T)=x}S_{0T}(\psi) \ .
\end{array}
\end{equation}

The last equality in the above display is due to the fact that we can always take a path $\psi(t)$, $0\leq t< \infty$ such that $\psi(t)=\psi(T)$ for $t\geq T$. Equation \eqref{Eq:StationaryMeasureExponentialForm} demonstrates a relation between the stationary measure and the ``action functional"
$S_{0T}(\psi)$ introduced in LDT.

\subsection{Local quasi--potential function as the solution to a variational problem and Hamilton--Jacobi equation.}
\label{Sec:LocalQuasiPotential:SubSec:LocalQuasi-potentialVariationalHJB}

From the above considerations, we can define a local quasi--potential function as

\begin{equation}\label{Eq:LocalQuasiPotentialVariationalProblem}
\phi_{\text{loc}}^{\text{QP}}(x;x_0)= \inf\limits_{T>0}\inf\limits_{\psi(0)=x_0, \psi(T)=x}S_{0T}(\psi) \ .
\end{equation}

Equations \eqref{Eq:StationaryMeasureExponentialForm} and
\eqref{Eq:LocalQuasiPotentialVariationalProblem} combined together gives
an exponential asymptotic for the stationary measure

\begin{equation}\label{Eq:DensityInvariantMeasureExponentialLocalQuasiPotential}
\rho^{\text{SS}}(x) \asymp \exp\left(-\dfrac{1}{\varepsilon}\phi^{\text{QP}}_{\text{loc}}(x; x_0)\right) \ ,
\end{equation}
which matches the identity \eqref{Eq:DensityInvariantMeasureExponentialQuasiPotential}.
This implies that in the case when there is only one stable attractor $O$ of the gradient system \eqref{Eq:GradientSystem},
the quasi--potential $\phi^{\text{QP}}(x)$ is given by the local quasi--potential
$\phi^{\text{QP}}_{\text{loc}}(x; x_0)$, which is the solution to a variational problem
\eqref{Eq:LocalQuasiPotentialVariationalProblem}.

\begin{remark}\label{Remark:LocalQuasiPotentialDependenceInitialPoint}
One may observe that our local quasi--potential function
$\phi^{\text{QP}}_{\text{loc}}(x;x_0)$ defined in \eqref{Eq:LocalQuasiPotentialVariationalProblem} depends on the initial condition $x_0$, while
 the stationary measure $\rho^{\text{SS}}(x)$ does not depend on $x_0$.
In fact, since \eqref{Eq:LDPEstimate} and \eqref{Eq:LDPPathSpaceDensity}
 describes the density function in path space of the process $x(t)$, for any $T>0$ we have
 $S_{0T}(\widehat{\psi})=0$ for $\widehat{\psi}(t)=x^{\text{GD}}(t)$ and $\widehat{\psi}(0)=x^{\text{GD}}(0)=x_0$.
However, when the loss function $f(x)$ is convex, all initial points $x_0\in \mathbb{R}^d$ are attracted by the gradient flow \eqref{Eq:GradientSystem}
 to the origin $O$. Combining these two effects, we can see that
 $\inf\limits_{T>0}\inf\limits_{\psi(0)=x_0, \psi(T)=x}S_{0T}(\psi)=\inf\limits_{T>0}\inf\limits_{\psi(0)=O, \psi(T)=x}S_{0T}(\psi)$, resulting in the fact that
 $\phi^{\text{QP}}_{\text{loc}}(x; x_0)=\phi^{\text{QP}}_{\text{loc}}(x; O)$ in the case when the loss function $f$ is convex.
However, in general when there are several different local minimum points of $f(x)$, one has to specify the local quasi--potential
with respect to the initial point $x_0$.
\end{remark}

\begin{remark}\label{Remark:ExchangingLimitOrder}
When obtaining
\eqref{Eq:StationaryMeasureExponentialForm} we have exchanged the limit order of $T\rightarrow\infty$ and $\varepsilon\rightarrow 0$.
This is because we have only one attractor $O$ of the gradient flow \eqref{Eq:GradientSystem}, so that in the long time
we do not expect to see transitions between different attractors
(the transitions between different attractors will be illustrated in the Section \ref{Sec:GlobalQuasiPotential}).
That being said, the asymptotic of $\rho(x,T|x_0,0)$ does not depend on the limit order $T\rightarrow \infty$ and $\varepsilon\rightarrow 0$, resulting
in the exponential form of the stationary measure \eqref{Eq:DensityInvariantMeasureExponentialLocalQuasiPotential}. See \cite{[Li-ZhouAIntegral]} for more details on this.
\end{remark}

We have seen that according to our general framework on the variational inference of SGD and its relation to the stationary measure,
SGD minimizes the local quasi--potential function $\phi^{\text{QP}}_{\text{loc}}(x;x_0)$ in the case when the loss function is convex.
The function $\phi^{\text{QP}}_{\text{loc}}(x;x_0)$ is shown to be
a solution to the variational problem \eqref{Eq:LocalQuasiPotentialVariationalProblem}.
In terms of implicit regularization, the quasi--potential $\phi^{\text{QP}}_{\text{loc}}(x;x_0)$
is related to diffusion matrix $D(x)$ in \eqref{Lemma:SDE-SGD:Eq:DiffusionMatrix} by solving the variational problem
\eqref{Eq:LocalQuasiPotentialVariationalProblem} via an explicit form of the action $S_{0T}(\psi)$.
According to LDT (see \cite{[FWbook1998]}, \cite{[FWbook2012]}, \cite{[Dembo-Zeitouni]}, \cite{[FWpaper1969]}
as well as Appendix \ref{Appendix:ElementsLDP}), when the
gradient flow \eqref{Eq:GradientSystem} has only one attractor $O$, the SGD diffusion equation \eqref{Lemma:SDE-SGD:Eq:SDE}
admits a LDT with the action functional (rate functional) given by the following explicit formula

\begin{equation}\label{Eq:ActionFunctional}
S_{0T}(\psi)=\left\{\begin{array}{l}
\displaystyle{\dfrac{1}{2}\int_0^T (\dot{\psi}_t+\nabla f(\psi_t))^T D^{-1}(\psi_t)(\dot{\psi}_t+\nabla f(\psi_t)) dt} \ ,
\\
\qquad \qquad \qquad \text{if } \psi_t \text{ is almost everywhere differentiable for}  \ t\in [0,T] \ ;
\\
+\infty \ ,
\\
\qquad \qquad \qquad \text{otherwise \ .}
\end{array}\right.
\end{equation}

Combining the formula \eqref{Eq:ActionFunctional} for the action functional $S_{0T}(\psi)$ and the variational formulation
of the quasi--potential, one obtains by using classical calculus of variations that the local quasi--potential function
$\phi^{\text{QP}}_{\text{loc}}(x;x_0)$ satisfies a partial differential equation of Hamilton--Jacobi type, which involves the diffusion matrix $D(x)$.
We have the following

\begin{theorem}\label{Thm:Hamilton-JacobiEqLocalQuasiPotential}
The local quasi--potential $\phi^{\text{QP}}_{\text{loc}}(x;x_0)$ is a solution to the Hamilton--Jacobi equation
\begin{equation}\label{Thm:Hamilton-JacobiEqLocalQuasiPotential:HJE}
\dfrac{1}{2}\left( \nabla \phi^{\text{QP}}_{\text{loc}}(x;x_0)\right)^T D(x) \nabla \phi^{\text{QP}}_{\text{loc}}(x;x_0)
-\nabla f(x)\cdot \nabla \phi^{\text{QP}}_{\text{loc}}(x;x_0)=0 \ ,
\end{equation}
with boundary condition $$\phi^{\text{QP}}_{\text{loc}}(O; x_0)=0 \ .$$
\end{theorem}

The proof is found in Appendix \ref{Appendix:HJB-Quasi-Potential}. The dependency of the Hamilton--Jacobi equation \eqref{Thm:Hamilton-JacobiEqLocalQuasiPotential:HJE} on the diffusion matrix
$D(x)$ can be viewed as a quantitative manifestation of the implicit regularization through the quasi--potential.
In particular, when $D(x)=I_d$, it is easy to see that $\phi^{\text{QP}}_{\text{loc}}(x;x_0)=2f(x)$ is a solution.
This justifies the prediction that for isotropic noise and only one minimizer, quasi--potential is just the original loss function
up to a multiplicative constant. Later in this Section we will provide an example with anisotropic noise covariance structure
and we will discuss properties of the quasi--potential in this case by making use of Theorem \ref{Thm:Hamilton-JacobiEqLocalQuasiPotential}.

\subsection{Escape properties from local minimum points in terms of the local quasi--potential.}
\label{Sec:LocalQuasiPotential:SubSec:EscapeLocalMinima}

Another remarkable feature of the local quasi--potential $\phi^{\text{QP}}_{\text{loc}}(x; x_0)$ is that it characterizes the escape
properties from local minimum points. As we have described in the first paragraph of this Section, the escape from sharp minima to flat minima
is a key feature which leads to good generalization (see \cite{[KeskarEtAlLargeBatchTraining]}). The LDT estimate
\eqref{Eq:LDPEstimate} provides a tool to obtain the exponential estimates of the exit probability and mean first exit time from the basin
of an attractor. To illustrate this, let us consider the set--up in our Section, where the loss function $f(x)$
is convex and admits only one single attractor $O$. Let $U$ be an open neighborhood of $O$ with boundary $\partial U$.
Let the process $x(t)$ start its motion from an initial point $x_0\in U$ and consider its first hitting time to $\partial U$:

\begin{equation}\label{Eq:FirstHittingTimeBoundaryU}
\tau(x_0, \partial U)=\inf\{t\geq 0: \ x(t)\in \partial U \ , \ x(0)=x_0\} \ .
\end{equation}

Suppose the SGD process $x(t)$ starts from some $x_0\in U$. Let $B(O, r)=\{x\in U, |x|\leq r\}$
be a closed ball around $O$ with radius $r>0$. Pick some small $\mu>0$ and let $\gamma=\partial B(O, \mu/2)$ and $\Gamma=\partial B(O, \mu)$ such that $\Gamma\subset U$. 
We introduce a sequence of Markov times $\tau_0\leq \sigma_0<\tau_1<\sigma_1<\tau_2<\sigma_2< ...$ in the following way:
let $\tau_0=0$ and $\sigma_n=\inf\{t>\tau_n: x(t)\in \Gamma\}$, $\tau_n=\inf\{t>\sigma_{n-1}: x(t)\in \gamma\cup \partial U\}$.
Consider the Markov chain $Z_n=x(\tau_n)$, $n=0,1,2,...$. The state space of $Z_n$, $n\geq 1$ is $\gamma\cup \partial U$.
Together with the hitting time $\tau(x_0, \partial U)$
we also define the one--step transition probability $P(x_0, \partial U)$ by
\begin{equation}\label{Eq:OneStepTransitionProbabilityBoundaryU}
P(x_0, \partial U)=\mathbf{P}\left(x(0)=x_0 \text{ and } Z_1\in \partial U\right) \ .
\end{equation}

Then we have the following Theorem characterizing the exponential asymptotic for the exit probability and
the mean exit time:

\begin{theorem}\label{Thm:ExponentialAsymptoticMeanExitTimeExitPosition}
Assume that the boundary $\partial U$ of the domain $U$ is smooth and $$-(\nabla f(x))^Tn(x)<0$$ for $x\in \partial U$, where
$n(x)$ is the exterior normal vector to the boundary of $U$, then for $x_0\in U$ we have the following two asymptotic
\begin{equation}\label{Thm:ExponentialAsymptoticMeanExitTimeExitPosition:Eq:ExponentialExitProbability}
\lim\limits_{\varepsilon\rightarrow 0, \mu \rightarrow 0}\varepsilon \ln P (x_0, \partial U)=-\min\limits_{x\in \partial U}\phi^{\text{QP}}_{\text{loc}}(x; O) \ ,
\end{equation}
and
\begin{equation}\label{Thm:ExponentialAsymptoticMeanExitTimeExitPosition:Eq:ExponentialExitTime}
\lim\limits_{\varepsilon\rightarrow 0}\varepsilon \ln \mathbf{E} \tau(x_0, \partial U)=\min\limits_{x\in \partial U}\phi^{\text{QP}}_{\text{loc}}(x; O) \ .
\end{equation}
\end{theorem}

The proof is found in Appendix \ref{Appendix:ElementsLDP}. Theorem \ref{Thm:ExponentialAsymptoticMeanExitTimeExitPosition} indicates that the probability of transition from $x_0\in U$
to $\partial U$ is given by the asymptotic
$$P(x_0, \partial U)\asymp \exp\left(-\dfrac{1}{\varepsilon}\min\limits_{x\in \partial U}\phi^{\text{QP}}_{\text{loc}}(x; O)\right) \ ,$$
and the mean exit time has exponential asymptotic
$$\mathbf{E}\tau(x_0, \partial U)\asymp \exp\left(\dfrac{1}{\varepsilon}\min\limits_{x\in \partial U}\phi^{\text{QP}}_{\text{loc}}(x; O)\right) \ .$$

\begin{remark}\label{Remark:ExitPosition}
It can also be shown that the local quasi--potential $\phi^{\text{QP}}_{\text{loc}}(x; O)$ is related to the first exit position.
In fact, assuming that $x^*$ is the only minimum point of $\phi^{\text{QP}}_{\text{loc}}(x; O)$ on $\partial U$,
then as $\varepsilon\rightarrow 0$,
the first exit position approaches the minimum point of $\phi^{\text{QP}}_{\text{loc}}(x; O)$
on $\partial U$, i.e.
\begin{equation}\label{Remark:ExitPosition:Eq:FirstExitPosition}
\lim\limits_{\varepsilon\rightarrow 0}x(\tau(x_0, \partial U))=x^*  \ , \ \text{ where } x^*=\arg\min\limits_{x\in \partial U}\phi^{\text{QP}}_{\text{loc}}(x; O) \ .
\end{equation}
\end{remark}

From here we can see that the escape properties of the process $x(t)$ from a local minimum point,
such as the exit probability, the mean escape time and even the first exit position, are related to
the quasi--potential $\phi^{\text{QP}}_{\text{loc}}(x; O)$.
This combined with Theorem \ref{Thm:Hamilton-JacobiEqLocalQuasiPotential} indicate that the choice of covariance structure
$D(x)$ affects the escape properties from local minimum points. By using these results,
the next example shows that in some cases, anisotropic noise helps the SGD process $x(t)$ escape faster from a local minimum point
than isotropic noise. This validates some of the predictions in \cite{[ZhanxingSGDAnisotropic]}.

\begin{example}\label{Example:AnisotropicNoiseFasterEscape}
Let $d=2$ and $x=(x_1, x_2)$, and consider the loss function $f(x)=x_1^2+x_2^2$. Let the neighborhood
$$U=\{(x_1,x_2): x_1^2+x_2^2<1\}$$
and initial condition $x_0\in U$.
Let $0<\mu<2$. Let
$D(x)=D^\mu(x)=\begin{pmatrix}\mu&0\\0&2-\mu\end{pmatrix}$. Notice that when $\mu=1$, $D^1(x)=I_2$ is isotropic, and when
$\mu\neq 1$, $D^\mu(x)$ is anisotropic.
It is easy to check by verifying
equation \eqref{Thm:Hamilton-JacobiEqLocalQuasiPotential:HJE} that for each $\mu\in (0,2)$ we have
$$\phi^{\text{QP}}_{\text{loc}}(x; O, \mu)=2\left[\dfrac{1}{\mu}x_1^2+\dfrac{1}{2-\mu}x_2^2\right] \ .$$
Thus
$$\begin{array}{ll}
\phi^{\text{QP}}_{\text{loc}}(x; O, \mu)& =2\left[\dfrac{1}{\mu}x_1^2+\dfrac{1}{2-\mu}x_2^2\right]
\\
& =2\left[\dfrac{1}{\mu}(x_1^2+x_2^2)+\left(\dfrac{1}{2-\mu}-\dfrac{1}{\mu}\right)x_2^2\right]
\\
& =2\left[\dfrac{1}{2-\mu}(x_1^2+x_2^2)+\left(\dfrac{1}{\mu}-\dfrac{1}{2-\mu}\right)x_1^2\right] \ .
\end{array}$$
\begin{itemize}
\item When $\mu=1$, the quasi--potential $\phi^{\text{QP}}_{\text{loc}}(x; O, 1)=2(x_1^2+x_2^2)=2$ on $x\in \partial U$.
Thus by \eqref{Thm:ExponentialAsymptoticMeanExitTimeExitPosition:Eq:ExponentialExitTime}
in Theorem \ref{Thm:ExponentialAsymptoticMeanExitTimeExitPosition} the mean exit time
$$\mathbf{E}\tau^1(x_0, \partial U)\asymp \exp\left(\dfrac{2}{\varepsilon}\right) \ .$$

\item When $1<\mu<2$, i.e. $\dfrac{1}{2-\mu}>\dfrac{1}{\mu}$, we have
$$\min\limits_{x\in \partial U}\phi^{\text{QP}}_{\text{loc}}(x; O, \mu)=\dfrac{2}{\mu}<2 \ .$$
Thus by \eqref{Thm:ExponentialAsymptoticMeanExitTimeExitPosition:Eq:ExponentialExitTime}
in Theorem \ref{Thm:ExponentialAsymptoticMeanExitTimeExitPosition} the mean exit time
$$\mathbf{E}\tau^\mu(x_0, \partial U)\asymp \exp\left(\dfrac{2}{\varepsilon\mu}\right)<\!\!<\mathbf{E}\tau^1(x_0, \partial U)$$
when $\varepsilon$ is small.

\item When $0<\mu<1$, i.e. $\dfrac{1}{\mu}>\dfrac{1}{2-\mu}$, we have
$$\min\limits_{x\in \partial U}\phi^{\text{QP}}_{\text{loc}}(x; O, \mu)=\dfrac{2}{2-\mu}<2 \ ;$$
Thus by \eqref{Thm:ExponentialAsymptoticMeanExitTimeExitPosition:Eq:ExponentialExitTime}
in Theorem \ref{Thm:ExponentialAsymptoticMeanExitTimeExitPosition} the mean exit time
$$\mathbf{E}\tau^\mu(x_0, \partial U)\asymp \exp\left(\dfrac{2}{\varepsilon(2-\mu)}\right)<\!\!<\mathbf{E}\tau^1(x_0, \partial U)$$
when $\varepsilon$ is small.
\end{itemize}
We have justified that in this case, anisotropic noise leads to faster escape then isotropic noise.
\end{example}

\section{Global quasi--potential: the case of multiple global minima and the stochastic dynamics of SGD.}
\label{Sec:GlobalQuasiPotential}

In the previous section we have introduced a local quasi--potential function $\phi_{\text{loc}}^{\text{QP}}(x)$ in
the case when the loss function $f(x)$ is convex and admits only one stable attractor $O$. In term of stationary measure,
from \eqref{Eq:StationaryMeasureExponentialForm} our
local quasi--potential function can be viewed as the following limit

$$\begin{array}{ll}
-\lim\limits_{\varepsilon\rightarrow 0}\varepsilon\ln \rho^{\text{SS}}(x) & = - \lim\limits_{\varepsilon\rightarrow 0}\lim\limits_{T\rightarrow\infty}\varepsilon\ln \rho(x,T|x_0,0)
\\
& =-\lim\limits_{T\rightarrow\infty}\lim\limits_{\varepsilon\rightarrow 0} \varepsilon\ln \rho(x,T|x_0,0)
\\
& =\phi_{\text{loc}}^{\text{QP}}(x; x_0) \ .
\end{array}$$

We have explained in Remark \ref{Remark:ExchangingLimitOrder}
that the exchange of limit order in the above demonstration is due to the fact that the loss function
$f(x)$ is convex and admits only one minimum point $O$.
In this section we consider the case when the loss function $f(x)$ is non--convex and possibly admits
several different local minimum points. Under this scenario, the exchange of limit order is not valid, and instead we have

\begin{equation}\label{Eq:GlobalQuasiPotentialStationaryMeasureExponentialForm}
\begin{array}{ll}
-\lim\limits_{\varepsilon\rightarrow 0}\varepsilon\ln \rho^{\text{SS}}(x) & = - \lim\limits_{\varepsilon\rightarrow 0}\lim\limits_{T\rightarrow\infty}\varepsilon\ln \rho(x,T|x_0,0)
\\
& \equiv \phi_{\text{glob}}^{\text{QP}}(x) \ .
\end{array}
\end{equation}

This defines the global quasi--potential function $\phi_{\text{glob}}^{\text{QP}}(x)$ or sometimes just abbreviated as the
quasi--potential function $\phi^{\text{QP}}(x)$. The asymptotic \eqref{Eq:GlobalQuasiPotentialStationaryMeasureExponentialForm}
indicates that we have the expression of the stationary measure $\rho^{\text{SS}}(x)$ just as we demonstrated in \eqref{Eq:DensityInvariantMeasureExponentialQuasiPotential}:

\begin{equation}\label{Eq:DensityInvariantMeasureExponentialQuasiPotential:SecGlobal}
\rho^{\text{SS}}(x) \asymp \exp\left(-\dfrac{1}{\varepsilon}\phi^{\text{QP}}(x)\right) \ .
\end{equation}

This relation between the quasi--potential $\phi^{\text{QP}}(x)$ and the stationary measure indicates that SGD minimizes
the quasi--potential function $\phi^{\text{QP}}(x)$. However, just such a definition did not provide useful information on
how to construct the quasi--potential $\phi^{\text{QP}}(x)$ and how it is related to the original loss function $f(x)$ and the
covariance structure $D(x)$.

In fact, the classical Large Deviations Theory (see \cite{[FWbook1998]}, \cite{[FWbook2012]}) provides a systematic yet rather complicated way to
construct the quasi--potential $\phi^{\text{QP}}(x)$ from its local version $\phi^{\text{QP}}_{\text{loc}}(x;x_0)$.
However, for the purpose of studying SGD we are mainly interested in the location of the local and global minimum points of
$\phi^{\text{QP}}(x)$ since these are the points that the SGD process will be finally trapped into.
Such a problem can be understood from the dynamics of SGD process \eqref{Lemma:SDE-SGD:Eq:SDE} via LDT Theorem \ref{Thm:ExponentialAsymptoticMeanExitTimeExitPosition}.
Let us illustrate this via a $2$--dimensional (i.e. in $\mathbb{R}^2$) example as follows.

\begin{example}\label{Example:TwoWellPotentialLossFunctionGlobalQuasipotential}
Consider the loss function $f(x)=f(x_1, x_2)$ defined in a piecewise way

\begin{equation}\label{Example:TwoWellPotentialLossFunctionGlobalQuasipotential:Eq:LossFunction}
f(x_1, x_2)=\left\{\begin{array}{lll}
(x_1+2)^2+x_2^2 & , & \text{ if }(x_1+2)^2+x_2^2\leq 1 \ ;
\\
1 & , & \text{ if }(x_1+2)^2+x_2^2>1 \text{ and } (x_1-2)^2+x_2^2>1\ ;
\\
(x_1-2)^2+x_2^2 & , & \text{ if }(x_1-2)^2+x_2^2\leq 1 \ .
\end{array}\right.
\end{equation}

The above defined loss function $f(x)$ has two local minimum points $O_1=(-2, 0)$ and $O_2=(2, 0)$.
The minimal values of $f(x)$ at $O_1$
and $O_2$ are both equal to $0$. The corresponding two basin of attractors
are $B_1=\{(x_1,x_2): (x_1+2)^2+x_2^2< 1\}$ and $B_2=\{(x_1, x_2): (x_1-2)^2+x_2^2< 1\}$, in which the function $f(x)$
increases from $0$ to $1$. Let us consider the piecewise defined gradient function

\begin{equation}\label{Example:TwoWellPotentialLossFunctionGlobalQuasipotential:Eq:GradientLoss}
\nabla f(x)=\left\{\begin{array}{lll}
(2(x_1+2), 2x_2) & , & \text{ if }(x_1+2)^2+x_2^2\leq 1 \ ;
\\
0 & , & \text{ if }(x_1+2)^2+x_2^2>1 \text{ and } (x_1-2)^2+x_2^2>1\ ;
\\
(2(x_1-2), 2x_2) & , & \text{ if }(x_1-2)^2+x_2^2\leq 1 \ .
\end{array}\right.
\end{equation}

Then we can consider the dynamics of the SGD process $x(t)$ in \eqref{Lemma:SDE-SGD:Eq:SDE}, with a chosen diffusion matrix
$D(x)$ such that $D(x)=\begin{pmatrix} \mu_1& 0\\ 0 & 2-\mu_1\end{pmatrix}$ when $x\in B_1$ and
$D(x)=\begin{pmatrix} \mu_2& 0\\ 0 & 2-\mu_2\end{pmatrix}$ when $x\in B_2$ for some $\mu_1, \mu_2\in (1,2)$ to be determined.
Let us also pick $D(x)=I_2$ to be the identity matrix when $x\not \in B_1 \text{ or } B_2$.

In a very similar fashion as Example \ref{Example:AnisotropicNoiseFasterEscape}, we can calculate the local quasi--potential
$\phi_{\text{loc}}^{\text{QP}}(x;x_0)$ with respect to $x_0=O_1 \text{ or } O_2$ within balls $B_1$ and $B_2$ as

$$\phi_{\text{loc}}^{\text{QP}}(x; O_1)=2\left[\dfrac{1}{\mu_1}(x_1+2)^2+\dfrac{1}{2-\mu_1}x_2^2\right] \ ,
\text{ when } (x_1+2)^2+x_2^2\leq 1 \ ,$$
and
$$\phi_{\text{loc}}^{\text{QP}}(x; O_2)=2\left[\dfrac{1}{\mu_2}(x_1-2)^2+\dfrac{1}{2-\mu_2}x_2^2\right] \ ,
\text{ when } (x_1-2)^2+x_2^2\leq 1 \ .$$
We use the same reasoning as in Example \ref{Example:AnisotropicNoiseFasterEscape}, this indicates that
$$\min\limits_{x\in \partial B_1}\phi_{\text{loc}}^{\text{QP}}(x; O_1)=\dfrac{2}{\mu_1} \ , \
\min\limits_{x\in \partial B_2}\phi_{\text{loc}}^{\text{QP}}(x; O_2)=\dfrac{2}{\mu_2} \ .$$

When the SGD process $x(t)$ in \eqref{Lemma:SDE-SGD:Eq:SDE} enters one of the basin of attractors, say $B_1$,
it will be attracted to $O_1$ by the gradient flow dynamics and fluctuate there due to the small noise term in \eqref{Lemma:SDE-SGD:Eq:SDE}.
The probability that it exits this basin is given by \eqref{Thm:ExponentialAsymptoticMeanExitTimeExitPosition:Eq:ExponentialExitProbability}
in Theorem \ref{Thm:ExponentialAsymptoticMeanExitTimeExitPosition}, that is
$$P(O_1, \partial B_1)\asymp \exp\left(-\dfrac{2}{\varepsilon \mu_1}\right) \ .$$
Once the process $x(t)$ reaches $\partial B_1$ with positive probability it will hit $\partial B_2$ and gets attracted to $O_2$. In this way,
the transition from $O_1$ to $O_2$ can be viewed as a Markov chain with transition probability
$$P(O_1, O_2)\asymp \exp\left(-\dfrac{2}{\varepsilon \mu_1}\right) \ .$$
Similarly, from $O_2$ a transition to $O_1$ may happen with probability
$$P(O_2, O_1)\asymp \exp\left(-\dfrac{2}{\varepsilon \mu_2}\right) \ .$$
Since $O_1$ and $O_2$ are two attractors of the SGD process with loss function $f(x)$ given by \eqref{Example:TwoWellPotentialLossFunctionGlobalQuasipotential:Eq:LossFunction}, the dynamics of SGD process spends most of its time near $O_1$
and $O_2$. Thus this dynamics can be viewed as a Markov chain on $O_1$ and $O_2$
with transition probabilities between them given by $P(O_1, O_2)$ and $P(O_2, O_1)$. This implies that the stationary measure $\rho^{SS}(x)$
concentrates on $O_1$ and $O_2$, with
$$
\rho^{SS}(O_1)= \dfrac{P(O_2, O_1)}{P(O_1, O_2)+P(O_2, O_1)}
\ , \
\rho^{SS}(O_2)= \dfrac{P(O_1, O_2)}{P(O_1, O_2)+P(O_2, O_1)}
\ .
$$
Set $\mu_1>\mu_2$. Then $\rho^{SS}(O_1)\rightarrow 0$ and $\rho^{SS}(O_2)\rightarrow 1$ as $\varepsilon\rightarrow 0$. This combined with \eqref{Eq:DensityInvariantMeasureExponentialQuasiPotential:SecGlobal} indicates that among the two local minimum points $O_1$
and $O_2$ of the loss function $f(x)$ in this example, the quasi--potential $\phi^{\text{QP}}(x)$ has local minima
at $O_1$ and $O_2$ with $\phi^{\text{QP}}(O_2)<\phi^{\text{QP}}(O_1)$. This indicates that the SGD process tends to select $O_2$
rather than $O_1$ as its final place to be trapped, even when the original loss function has same values on both $O_1$ and $O_2$.
This ``selection of specific local minimum point" can be viewed as a regularization induced by the anisotropic noise covariance structure.
\end{example}

\section{Numerical Experiments.}
\label{Sec:Numerics}

We have performed numerical experiment for Example 3.1. Here we pick
$\varepsilon=0.1$ and stepsize $0.01$. In Figure 1 (a)--(d) the number of iterations is equal to $140000$. Figure 1 (a) and (c) (blue) are for isotropic noise $\mu=1$ and Figure 1 (b) and (d)
are for anisotropic noise $\mu=1.9999$. Figure 1 (a) and (b) show the evolution of the radial processes $R(t)=\sqrt{x_1^2(t)+x_2^2(t)}$ for the isotropic and anisotropic cases. It is observed that at number of iteration $140000$
the anisotropic process already exits the basin of attractor $U=\{(x_1,x_2): x_1^2+x_2^2<1\}$
while the isotropic process does not exit. Moreover, comparing Figure 1 (c) and (d) we
see that the process with anisotropic noise tends to exit along the horizontal direction, while the
process with isotropic noise tend to distribute more evenly within the basin of attraction.

\begin{figure}[H]
\centering
\includegraphics[height=13cm, width=15cm]{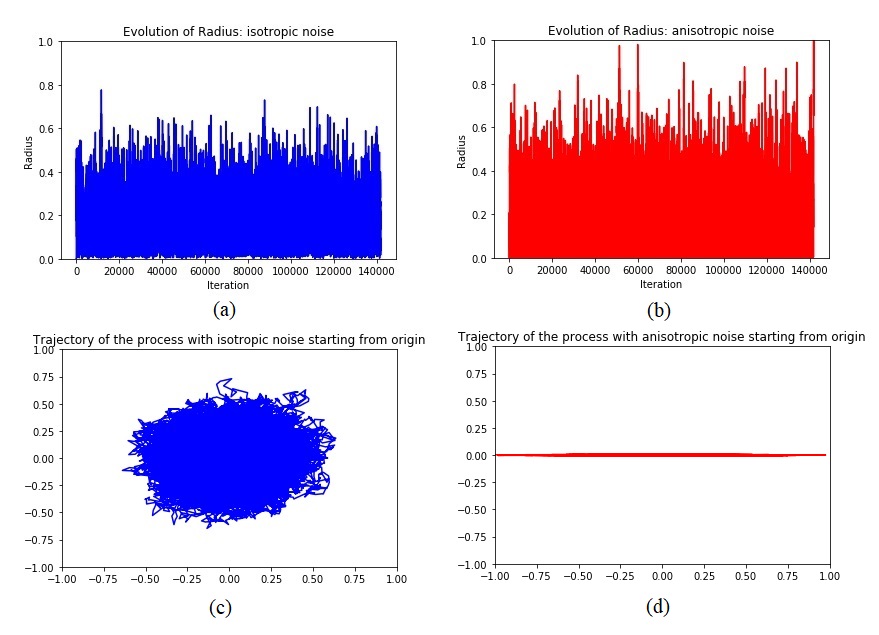}
\caption{Numerical Experiment for Example 3.1.}
\end{figure}

We have also performed numerical experiment for Example 4.1. Here we pick $\varepsilon=0.2$ and stepsize $0.01$.
We pick $\mu_1=1.9999$ and $\mu_2=1.0001$. In both Figure 2(a) and Figure 2(b) the number of iterations are equal to
$22000$. However, in Figure 2(a) the process starts from $O_1=(-2, 0)$ and in Figure 2(b) the process starts from
$O_2=(2,0)$. It is observed that compared to $O_1$, SGD process tends to select the minimum $O_2$ as its final
place to be trapped. Indeed, in Figure 2(a) when the SGD process starts from $O_1$, it will exit the basin of attractor
$B_1$ and gets trapped to $B_2$. This is because within the two basins of attraction the noise covariance structures are different:
$\mu_1=1.9999$ corresponds to highly anisotropic noise and $\mu_2=1.0001$ corresponds to almost isotropic noise.
In Figure 2(b) we see that when the SGD process starts from $O_2$, it tends to stay within this basin of attraction $B_2$.
Although the loss functions have the same depth at both local minima $O_1$ or $O_2$, the selection of the minimum point $O_2$
by the SGD process can be viewed as a result of implicit regularization due to different noise covariance structures
within different basins of attraction.

\begin{figure}[H]
\centering
\includegraphics[height=7cm, width=15cm]{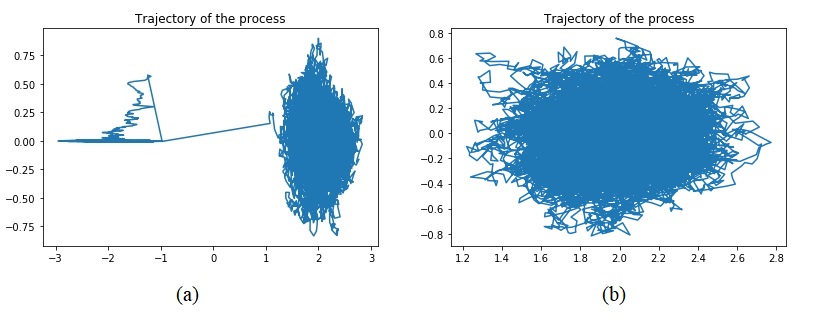}
\caption{Numerical Experiment for Example 4.1.}
\end{figure}

\section{Conclusion and Future Directions.}\label{Sec:Conclusion}

We proposed in this work a unified way to
enhance the understanding of the connection between SGD's noise covariance structure and its
selection of specifically favored local minimum points.
We have introduced a new potential function named the \textit{quasi--potential} and
we interpret the variational inference of SGD as minimizing the quasi--potential function.
LDT and classical calculus of variations enable us to find
a relation between the quasi--potential and the noise covariance structure (the diffusion matrix in SGD).
This relation is concretely and quantitatively built upon a partial differential equation of Hamilton--Jacobi type.
By solving the Hamilton--Jacobi equation in specific cases, we find in an example that
anisotropic noise leads to faster escape than isotropic noise.
We also propose to consider the stochastic dynamics of SGD associated with the quasi--potential.
In this aspect, the mechanism of ``implicit regularization" was explained through a Markov chain
between local minimum points of the quasi--potential.
We understand that SGD selects its favored local minimum points through performing a Markov chain
between different local minima, and the behavior of this Markov chain is related to SGD's noise covariance structure.
Thus we provide here a quantitative way to understand the phenomenon of ``implicit regularization"
by proposing the quasi--potential and relate it to the noise covariance structure via partial differential equation
and stochastic dynamics.

As for a future direction, a natural question to ask is how this general framework can be carried to particular machine learning models,
such as neural networks. Indeed it has been found that highly anisotropic noise is common in the SGD training of deep neural networks
(see \cite{[Chaudhari-SoattoSGDVariational]}, \cite{[ZhanxingSGDAnisotropic]}, \cite{[HofmannEtAlEscapeSaddleSGD]}).
Thus it would be interesting to see if the anisotropic noise structure affects the escape and generalization properties of SGD
via our proposed quasi--potential function. In particular, by looking at the properties of the Hamilton--Jacobi equation
\eqref{Thm:Hamilton-JacobiEqLocalQuasiPotential:HJE} that involves an anisotropic diffusion matrix $D(x)$, we may have a better interpretation
of ``implicit regularization" in concrete deep learning models. This is left to future work.

\bibliographystyle{plain}
\bibliography{bibliography}

\newpage

%Appendix to the quasi-potential implicit regularizer paper

\begin{appendix}
\section{Elements of Large Deviations Theory (LDT).}
\label{Appendix:ElementsLDP}

The Large Deviations Theory (LDT) is a mathematical theory that characterizes the probability
of very rare events. These rare events happen with exponentially small probabilities, and LDT quantifies
the asymptotic of the exponent. To illustrate this, let us consider the SGD process \eqref{Lemma:SDE-SGD:Eq:SDE}
that we have introduced in Lemma \ref{Lemma:SDE-SGD}:

\begin{equation}\label{Appendix:ElementsLDP:Eq:SGD-SDE}
dx(t)=-\nabla f(x(t))dt+\sqrt{\varepsilon} \Sigma(x(t))dW(t) \ , \ x(0)=x_0\in \mathbb{R}^d \ .
\end{equation}

When the parameter $\varepsilon>0$ is small, the trajectory of $x(t)$ in finite time stays $\mathcal{O}(\sqrt{\varepsilon})$--close
to the Gradient Descent (GD) flow characterized by the deterministic equation

\begin{equation}\label{Appendix:ElementsLDP:Eq:GradientSystem}
\dfrac{dx^{\text{GD}}(t)}{dt}=-\nabla f(x^{\text{GD}}(t)) \ .
\end{equation}

The above deterministic GD flow attracts the process $x^{\text{GD}}(t)$ to critical points
(such as local minimum points) of the loss function
$f(x)$. Thus for small $\varepsilon>0$ the SGD process $x(t)$ should also be attracted to a
neighborhood of some critical point
of the loss function $f(x)$, as the two processes stay $\mathcal{O}(\sqrt{\varepsilon})$--close.
However, this $\mathcal{O}(\sqrt{\varepsilon})$--difference is caused by the random fluctuation term
$\sqrt{\varepsilon} \Sigma(x(t))dW(t)$ in \eqref{Appendix:ElementsLDP:Eq:SGD-SDE}.
Such a random fluctuation term may cause not only an $\mathcal{O}(\sqrt{\varepsilon})$--difference between the processes
$x^{\text{GD}}(t)$ and $x(t)$, but also large fluctuations of $x(t)$ from $x^{\text{GD}}(t)$ due to its random nature.
Since the randomness is small with a scale of $\sqrt{\varepsilon}$, the large scale fluctuations
happen with very small probability, and thus they are named ``rare events".

Mathematically, these large fluctuations of $x(t)$ from $x^{\text{GD}}(t)$ can be characterized in a functional way.
That is to say, we can take a reference path $\psi(t)$, $0\leq t \leq T$ and some $\delta>0$. Then we consider the probability
that the trajectory of $x(t)$, $0\leq t \leq T$ stays in a $\delta$--neighborhood of the reference path $\psi(t)$:

\begin{equation}\label{Appendix:ElementsLDP:Eq:LDPEstimate}
\mathbf{P}\left(\sup\limits_{0\leq t\leq T}|x(t)-\psi(t)|\leq \delta\right)\asymp \exp\left(-\dfrac{1}{\varepsilon}S_{0T}(\psi)\right)  \ .
\end{equation}

The asymptotic \eqref{Appendix:ElementsLDP:Eq:LDPEstimate} is in terms of a logarithmic equivalence and the functional $S_{0T}(\psi)$
on the exponent of \eqref{Appendix:ElementsLDP:Eq:LDPEstimate} is called the \textit{action functional} (or \textit{rate functional}).
The functional $S_{0T}(\psi)$ characterizes the difficulty of any fluctuation of $x(t)$ deviating from $x^{\text{GD}}(t)$
that happens close to the given trajectory $\psi(t)$, $0\leq t \leq T$.
Usually, the trajectory $\psi(t)$ is taken from $\mathbf{C}_{[0, T]}(\mathbb{R}^d)$: the family
of continuous functions on $[0,T]$ that maps into $\mathbb{R}^d$.
For example, if the trajectory $\psi(t)$ is taken as the GD
flow $x^{\text{GD}}(t)$ in \eqref{Appendix:ElementsLDP:Eq:GradientSystem}, then
we have $S_{0T}(\psi)=0$. This indicates that as $\varepsilon\rightarrow 0$,
most of the possible trajectories of $x(t)$ are concentrated around the GD flow $x^{\text{GD}}(t)$.
The random noise term $\sqrt{\varepsilon}\Sigma(x(t))dW(t)$ may still cause a small portion
of the trajectories that are not around the GD flow $x^{\text{GD}}(t)$, and these are characterized by trajectories
$\psi(t)$ such that $S_{0T}(\psi)>0$.

These above heuristics can be formally stated in the following LDT Theorem.

\begin{theorem}\label{Appendix:ElementsLDP:Thm:LargeDeviations}(Large Deviations Theory)
Let $\varepsilon>0$. Then the family of processes $x(t)$, $0\leq t \leq T$, $x(0)=x_0\in \mathbb{R}^d$ satisfies a Large Deviations Theory (LDT)
with rate functional (action functional) $S_{0T}(\psi)$ in the space $\mathbf{C}_{[0,T]}(\mathbb{R}^d)$ with the following properties:
\begin{itemize}
\item[(1)] Given each $s>0$ the set $\Psi(s)=\{\psi: S_{0T}(\psi)\leq s\}$ is compact in $\mathbf{C}_{[0,T]}(\mathbb{R}^d)$;
\item[(2)] For any $\delta>0$, any $\gamma>0$ and any $s_0>0$ there exists a positive $\varepsilon_0>0$ such that for any $0<\varepsilon<\varepsilon_0$
we have
\begin{equation}\label{Appendix:ElementsLDP:Thm:LargeDeviations:Eq:LDPLowerBound}
\mathbf{P}\left(\sup\limits_{0\leq t\leq T}|x(t)-\psi(t)|<\delta\right)\geq \exp\left(-\dfrac{1}{\varepsilon}(S_{0T}(\psi)+\gamma)\right) \ ,
\end{equation}
where the given path $\psi\in \mathbf{C}_{[0,T]}(\mathbb{R}^d)$ is such that $\psi(0)=x_0$ and $S_{0T}(\psi)\leq s_0$;
\item[(3)] For any $\delta>0$, any $\gamma>0$ and any $s_0>0$ there exists a positive $\varepsilon_0>0$ such that for any $0<\varepsilon<\varepsilon_0$
and any $0<s<s_0$ we have
\begin{equation}\label{Appendix:ElementsLDP:Thm:LargeDeviations:Eq:LDPUpperBound}
\mathbf{P}\left(\inf\limits_{\psi\in \Psi_{x_0}(s)}\sup\limits_{0\leq t\leq T}|x(t)-\psi(t)|\geq \delta\right)\leq \exp\left(-\dfrac{1}{\varepsilon}(s-\gamma)\right) \ ,
\end{equation}
where $\Psi_{x_0}(s)=\{\psi\in \mathbf{C}_{[0,T]}(\mathbb{R}^d): \psi(0)=x_0 \ , \ S_{0T}(\psi)\leq s\}$.
\end{itemize}
Moreover, the SGD process $x(t)$ in \eqref{Appendix:ElementsLDP:Eq:SGD-SDE} corresponds to an analytically explicit form
of the rate functional (action functional) as
\begin{equation}\label{Appendix:ElementsLDP:Thm:LargeDeviations:Eq:ActionFunctional}
S_{0T}(\psi)=\left\{\begin{array}{l}
\displaystyle{\dfrac{1}{2}\int_0^T (\dot{\psi}_t+\nabla f(\psi_t))^T D^{-1}(\psi_t)(\dot{\psi}_t+\nabla f(\psi_t)) dt} \ ,
\\
\qquad \qquad \qquad \text{if } \psi_t \text{ is differentiable for almost every}  \ t\in [0,T] \ ;
\\
+\infty \ ,
\\
\qquad \qquad \qquad \text{otherwise \ .}
\end{array}\right.
\end{equation}
Here the covariance matrix $D(x)=\Sigma(x)\Sigma^T(x)$ is assumed to be invertible.
\end{theorem}

The proof of Theorem \ref{Appendix:ElementsLDP:Thm:LargeDeviations} can be found in several monographs dedicated to LDT. See
for example, \cite[\S 5.2 and \S 5.3]{[FWbook2012]}, \cite[\S 5.6]{[Dembo-Zeitouni]}. Applications of this Theorem can be found in e.g. 
\cite{[Freidlin-HuLandau-Lifschitz]}, \cite{[Hu-Metastability-Nearly-Elastic]}, \cite{[Hu-Lucas]}.

There are many ways to apply the above LDT Theorem. In particular, it can be used to prove Theorem
\ref{Thm:ExponentialAsymptoticMeanExitTimeExitPosition} in Section \ref{Sec:LocalQuasiPotential} of this paper.
Let us first explain the heuristics. The set--up of Theorem
\ref{Thm:ExponentialAsymptoticMeanExitTimeExitPosition} assumes that the loss function $f(x)$ admits
only one global minimum point $O$. Consider an open neighborhood $U$ of the origin $O$ with boundary $\partial U$.
Theorem \ref{Thm:ExponentialAsymptoticMeanExitTimeExitPosition} also assumes
that $-(\nabla f(x))^Tn(x)<0$ for all $x\in \partial U$, where $n(x)$ is the exterior normal vector
to the boundary of $U$. These conditions ensure that all points in $U$ are attracted by the GD flow \eqref{Appendix:ElementsLDP:Eq:GradientSystem}
to the origin $O$. By our heuristic argument before we state Theorem \ref{Appendix:ElementsLDP:Thm:LargeDeviations},
we see that if we start the process $x(t)$ from some point $x(0)=x_0\in U$, then as the parameter $\varepsilon>0$ is small the process $x(t)$
will also be attracted close to the origin $O$, since $x(t)$ and $x^{\text{GD}}(t)$ are $\mathcal{O}(\sqrt{\varepsilon})$--close to each other.
In this case, random noise term $\sqrt{\varepsilon}\Sigma(x(t))dW(t)$ induces ``escape" of the process $x(t)$ from the origin $O$ to the boundary $\partial U$.
The ``rare events" here correspond to an escape from $O$ to $\partial U$, and it has exponentially small probability.
To characterize these rare events, we shall make use of the functional characterization that we have explained before.
That is, we can take a deterministic continuous path $\psi\in \mathbf{C}_{[0,T]}(\mathbb{R}^d)$
that connects $O$ to $\partial U$: $\psi(0)=O$ and $\psi(T)\in \partial U$. According to the ansatz \eqref{Appendix:ElementsLDP:Eq:LDPEstimate},
the mathematical content of which is Theorem \ref{Appendix:ElementsLDP:Thm:LargeDeviations}, the probability of an ``escape"
along this given path $\psi$ has the exponential asymptotic $\exp\left(-\dfrac{1}{\varepsilon}S_{0T}(\psi)\right)$.
As we consider all possible ``escapes" from $O$ to $\partial U$, the one with highest probability corresponds to the first exit.
This probability should be given by the variational infimum of the functional $S_{0T}(\psi)$:
\begin{equation}\label{Appendix:ElementsLDP:Thm:LargeDeviations:Eq:LocalQuasiPotentialVariationalProblem}
\min\limits_{x\in \partial U}\phi_{\text{loc}}^{\text{QP}}(x;O)=
\min\limits_{x\in \partial U}\inf\limits_{T>0}\inf\limits_{\psi(0)=O, \psi(T)=x}S_{0T}(\psi) \ .
\end{equation}
Such a heuristic argument explains
\eqref{Thm:ExponentialAsymptoticMeanExitTimeExitPosition:Eq:ExponentialExitProbability}
in Theorem \ref{Thm:ExponentialAsymptoticMeanExitTimeExitPosition}
and \eqref{Remark:ExitPosition:Eq:FirstExitPosition} in Remark \ref{Remark:ExitPosition}. To understand
\eqref{Thm:ExponentialAsymptoticMeanExitTimeExitPosition:Eq:ExponentialExitTime}, we can view each escape as
a Bernoulli trial with success probability given by
\eqref{Thm:ExponentialAsymptoticMeanExitTimeExitPosition:Eq:ExponentialExitProbability}. Thus
the number of trials, i.e., the first exit time is inverse proportional to the probability given by
\eqref{Thm:ExponentialAsymptoticMeanExitTimeExitPosition:Eq:ExponentialExitProbability}.

A rigorous but very long and technical mathematical proof of Theorem \ref{Thm:ExponentialAsymptoticMeanExitTimeExitPosition} can be found in
\cite[\S 4.2, Theorem 4.2.1, \S 4.4, Theorem 4.4.1]{[FWbook2012]} or \cite[\S 5.7]{[Dembo-Zeitouni]}. For a more original proof see \cite{[FWpaper1969]}.
A proof of the claim in Remark \ref{Remark:ExitPosition} is found in \cite[\S 4.2, Theorem 4.2.1]{[FWbook2012]}.
Below we provide a sketch of this proof just for the reader's convenience. Our proof will be based on
the LDT Theorem \ref{Appendix:ElementsLDP:Thm:LargeDeviations} and a few
 technical lemmas (Lemmas 4.2.1, 4.2.2 and 4.2.3 in \cite{[FWbook2012]}) that we will just refer to
 \cite{[FWbook2012]}.

\

\textbf{Sketch of the Proof of Theorem \ref{Thm:ExponentialAsymptoticMeanExitTimeExitPosition}.}

We will make use of the stopping time $\tau(x_0, \partial U)$ and the
sequence of Markov times $\tau_0\leq \sigma_0<\tau_1<\sigma_1<\tau_2<\sigma_2<...$ and the Markov chain $Z_n$
that we introduced in Section \ref{Sec:LocalQuasiPotential:SubSec:EscapeLocalMinima}.
We would like to make use of the probability upper and lower estimations
\eqref{Appendix:ElementsLDP:Thm:LargeDeviations:Eq:LDPLowerBound}, \eqref{Appendix:ElementsLDP:Thm:LargeDeviations:Eq:LDPUpperBound}
at the level of sample paths. To this end, the key point in the proof is to construct
various trajectories $\psi(t)$, $0\leq t\leq T$ on which $S_{0T}(\psi)$ can be controlled from
$V_0\equiv \min\limits_{x\in \partial U}\phi_{\text{loc}}^{\text{QP}}(x;O)>0$ in
\eqref{Appendix:ElementsLDP:Thm:LargeDeviations:Eq:LocalQuasiPotentialVariationalProblem}.
To illustrate this, let us choose an arbitrarily small $d>0$. For $r>0$ small enough we
define $B(O, r)=\{x\in U: |x-O|\leq r\}$ to be a closed ball around $O$ with radius $r>0$.

\textbf{Step 1. Proof of \eqref{Thm:ExponentialAsymptoticMeanExitTimeExitPosition:Eq:ExponentialExitProbability}.}

We choose $\mu, \delta, T_1, T_2>0$ such that the following conditions are satisfied: first, since $O$
attracts all of $U$ and we have $-(\nabla f(x))^T n(x)<0$ for $x\in \partial U$,
all trajectories of the unperturbed system, that is the flow $x^{\text{GD}}(t)$ in \eqref{Appendix:ElementsLDP:Eq:GradientSystem},
starting at points $x_0\in U\cup \partial U$, hits
$B(O, \mu/2)$ before time $T_1$ and after time $T_1$ they do not leave $B(O, \mu/2)$;
secondly, for any $x\in B(O, \mu/2)$ we pick a function $\psi^x(t)$, $0\leq t\leq T_2=T(x)$ such that
$\psi^x(0)=x$ and there exists $T_3>0$ such that $\psi^x(t)\in B(O, \mu/2)$ for $0\leq t \leq T_3$, with
$\psi^x(T_3)=O$ and $S_{0T_3}(\psi^x)\leq 0.1d$. This is guaranteed based on Lemma 4.2.3 in \cite{[FWbook2012]}.
Starting from $\psi^x(T_3)=O$, we consider another path $\widehat{\psi}^x(0)=O$ and
$\widehat{\psi}^x(T_4)$ reaches the exterior of the $\delta$--neighborhood of $U$, does not hit $B(O, \mu/2)$ after exit from $U$
and $S_{0T_4}(\widehat{\psi}^x)<V_0+0.5d$. This can be done by letting $\widehat{\psi}^x$ be close to the trajectory
that minimizes the action functional. Let this path $\widehat{\psi}^x$ last hit $\Gamma$ at $x_1$ and we truncate
the part of this trajectory $\widehat{\psi}^x$ from $x_1$ to its final point $\widehat{\psi}^x(T_4)$.
This gives us a new path $\widetilde{\psi}^x(t)$, $0\leq t\leq T_5$ such that $\widetilde{\psi}^x(0)=x_1$,
$\widetilde{\psi}^x(T_5)$ reaches the exterior of the $\delta$--neighborhood of $U$, does not hit $B(O, \mu/2)$ after exit from $U$
and $S_{0T_5}(\widetilde{\psi}^x)<V_0+0.5d$. We then connect $O$ with $x_1$ by taking a path
$\psi^x(t)$, $T_3\leq t\leq T_3+T_6$ such that $\psi^x(T_3)=O$ and $\psi^x(T_3+T_6)=x_1$ with
$S_{T_3, T_3+T_6}(\psi^x)\leq 0.1d$. This is again based on Lemma 4.2.3 in \cite{[FWbook2012]}.
After that, we set $\psi^x(t)=\widetilde{\psi}^x(t-T_3-T_6)$ for $T_3+T_6\leq t\leq T_3+T_6+T_5$. Set $T_2=T(x)=T_3+T_6+T_5$ and we have
$S_{0T}(\psi)<V_0+0.5d+0.1d+0.1d=V_0+0.7d$.
With these construction at hand we make use of \eqref{Appendix:ElementsLDP:Thm:LargeDeviations:Eq:LDPLowerBound}
with $\gamma=0.3d$ to see that for $y\in B(O, \mu/2)$ we have
\begin{equation}\label{Thm:ExponentialAsymptoticMeanExitTimeExitPosition:Eq:ExitProbabilityLowerBound-path}
\mathbf{P}_{y}\left(\sup\limits_{0\leq t\leq T(y)}|x(t)-\psi^{y}(t)|<\delta\right)\geq
\exp\left(-\dfrac{1}{\varepsilon}\left(S_{0T(y)}(\psi^{y})+\gamma\right)\right)
\geq
\exp\left(-\dfrac{1}{\varepsilon}\left(V_0+d\right)\right)
\end{equation}
whenever $\varepsilon>0$ is sufficiently small. Since $\psi^{y}(T(y))$ reaches
the exterior of the $\delta$--neighborhood of $U$ at time $T=T(y)$, the condition
$\sup\limits_{0\leq t\leq T(y)}|x(t)-\psi^{y}(t)|<\delta$ ensures that $x(T(y))$ already exits $U$.
Moreover, starting from $x(0)=y$ the path $x(t)$ with $\sup\limits_{0\leq t\leq T(y)}|x(t)-\psi^{y}(t)|<\delta$
and $\delta<\mu/2$ we see that $x(t)$ hits $\Gamma$ first and then without returning to $\gamma$ it hits $\partial U$.
We notice that starting from $x(0)=x_0\in U$ the first part of the above construction ensures
that in time $T_1$ the deterministic trajectory $x^{\text{GD}}(t)$
is attracted to $B(O, \mu/2)$, and for small $\varepsilon>0$ the
 trajectory of $x(t)$ converge in probability to the deterministic trajectory $x^{\text{GD}}(t)$ (Lemma 4.2.1 in \cite{[FWbook2012]}).
These considerations together with \eqref{Thm:ExponentialAsymptoticMeanExitTimeExitPosition:Eq:ExitProbabilityLowerBound-path} imply that
\begin{equation}\label{Thm:ExponentialAsymptoticMeanExitTimeExitPosition:Eq:ExitProbabilityLowerBound}
P(x_0, \partial U)=\mathbf{P}\left(x(0)=x_0 \text{ and } Z_1\in \partial U\right)
\geq
\exp\left(-\dfrac{1}{\varepsilon}\left(V_0+d\right)\right) \ ,
\end{equation}
where $Z_n$ is the Markov chain defined in Section \ref{Sec:LocalQuasiPotential:SubSec:EscapeLocalMinima}. This is the lower bound for
\eqref{Thm:ExponentialAsymptoticMeanExitTimeExitPosition:Eq:ExponentialExitProbability}.

To obtain an upper bound for \eqref{Thm:ExponentialAsymptoticMeanExitTimeExitPosition:Eq:ExponentialExitProbability}, we make use of
\eqref{Appendix:ElementsLDP:Thm:LargeDeviations:Eq:LDPUpperBound} in the LDT Theorem \ref{Appendix:ElementsLDP:Thm:LargeDeviations}.
We notice first that for a fixed $T>0$ we have
\begin{equation}\label{Thm:ExponentialAsymptoticMeanExitTimeExitPosition:Eq:ExitProbabilityUpperBound-Truncation}
\begin{array}{ll}
P(x_0, \partial U) & \leq \max\limits_{y\in \Gamma}\mathbf{P}_y(\tau_1=\tau(x_0, \partial U))
\\
& \leq \max\limits_{y\in \Gamma}\left[\mathbf{P}_y(\tau_1=\tau(x_0, \partial U)<T)+\mathbf{P}_y(\tau_1=\tau(x_0, \partial U)\geq T)\right] \ .
\end{array}
\end{equation}

By Lemma 4.2.2 in \cite{[FWbook2012]}, $T$ can be chosen so large that we have an estimate for the second probability

\begin{equation}\label{Thm:ExponentialAsymptoticMeanExitTimeExitPosition:Eq:ExitProbabilityUpperBound-EasyPart}
\mathbf{P}_y(\tau_1=\tau(x_0, \partial U)\geq T)\leq \dfrac{1}{2}\exp\left(-\dfrac{1}{\varepsilon}(V_0-0.9d)\right) \ .
\end{equation}

We then notice that the trajectories of $x(t)$, $0\leq t\leq T$ for which $\tau(x_0, \partial U)=\tau_1<T$ are at a positive
distance $\delta>0$ from the set of functions $\Psi_y(V_0-0.5d)=\{\psi\in \mathbf{C}_{[0,T]}(\mathbb{R}^d): \psi(0)=y, S_{0T}(\psi)\leq V_0-0.5d\}$
provided that $d>0$ is arbitrary and $\mu>0$ is small. Pick $\gamma=0.4d>0$ and apply \eqref{Appendix:ElementsLDP:Thm:LargeDeviations:Eq:LDPUpperBound}
in the LDT Theorem \ref{Appendix:ElementsLDP:Thm:LargeDeviations} with $s=V_0-0.5d$, we see that

\begin{equation}\label{Thm:ExponentialAsymptoticMeanExitTimeExitPosition:Eq:ExitProbabilityUpperBound-HardPart}
\mathbf{P}_y(\tau_1=\tau(x_0, \partial U)< T)\leq \exp\left(-\dfrac{1}{\varepsilon}(V_0-0.9d)\right) \ .
\end{equation}

Combining \eqref{Thm:ExponentialAsymptoticMeanExitTimeExitPosition:Eq:ExitProbabilityUpperBound-Truncation},
\eqref{Thm:ExponentialAsymptoticMeanExitTimeExitPosition:Eq:ExitProbabilityUpperBound-EasyPart},
\eqref{Thm:ExponentialAsymptoticMeanExitTimeExitPosition:Eq:ExitProbabilityUpperBound-HardPart}
we see that we have the upper bound for \eqref{Thm:ExponentialAsymptoticMeanExitTimeExitPosition:Eq:ExponentialExitProbability}.
As $d$ is small and $\varepsilon>0$ is small we have

\begin{equation}\label{Thm:ExponentialAsymptoticMeanExitTimeExitPosition:Eq:ExitProbabilityUpperBound}
P(x_0, \partial U)\leq
\exp\left(-\dfrac{1}{\varepsilon}\left(V_0-d\right)\right) \ .
\end{equation}

The estimates \eqref{Thm:ExponentialAsymptoticMeanExitTimeExitPosition:Eq:ExitProbabilityLowerBound}, \eqref{Thm:ExponentialAsymptoticMeanExitTimeExitPosition:Eq:ExitProbabilityUpperBound} concludes the proof of
\eqref{Thm:ExponentialAsymptoticMeanExitTimeExitPosition:Eq:ExponentialExitProbability} in Theorem \ref{Thm:ExponentialAsymptoticMeanExitTimeExitPosition}.

\textbf{Step 2. Proof of \eqref{Thm:ExponentialAsymptoticMeanExitTimeExitPosition:Eq:ExponentialExitTime}.}

The lower and upper estimates \eqref{Thm:ExponentialAsymptoticMeanExitTimeExitPosition:Eq:ExitProbabilityLowerBound},
\eqref{Thm:ExponentialAsymptoticMeanExitTimeExitPosition:Eq:ExitProbabilityUpperBound} can actually be identified as the success probability
estimates for the Bernoulli trials which are attempts of touching $\partial U$ by the process $x(t)$.
To obtain the exit time estimate \eqref{Thm:ExponentialAsymptoticMeanExitTimeExitPosition:Eq:ExponentialExitTime}, we can just turn
the above ``success probability estimates of Bernoulli trials" into ``estimates of the number of trials until the first success"
via the Markov property of the chain $Z_n$. Indeed by our constructions in Step 1 of this proof which lead to \eqref{Thm:ExponentialAsymptoticMeanExitTimeExitPosition:Eq:ExitProbabilityLowerBound-path} we see that
for $y\in B(O, \mu/2)$ we have

$$\mathbf{P}_y\left(\tau(y, \partial U)<T_2\right)\geq \exp\left(-\dfrac{1}{\varepsilon}(V_0+d)\right)  \ .$$

Starting from $x_0\in U$ we set the first entrance time $\mathcal{T}=\min\{t: x(t)\in B(O, \mu/2)\}$.
Then by Markov property of $x(t)$ we have
$$\mathbf{P}_{x_0}(\tau(x_0, \partial U)<T_1+T_2)\geq \mathbf{P}_{x_0}(\mathcal{T}<T_1)\mathbf{P}_{x(\mathcal{T})}(\tau(x(\mathcal{T}))<T_2)\geq
\dfrac{1}{2}\exp\left(-\dfrac{1}{\varepsilon}(V_0+d)\right) \ .$$

Therefore by Markov property of $x(t)$ we have

\begin{equation}\label{Thm:ExponentialAsymptoticMeanExitTimeExitPosition:Eq:ExponentialExitTimeUpperBound}
\begin{array}{ll}
\mathbf{E} \tau(x_0, \partial U) & = \displaystyle{\sum\limits_{n=1}^\infty (n+1)(T_1+T_2)\mathbf{P}_{x_0}\left( n(T_1+T_2)<\tau(x_0, \partial U)\leq (n+1)(T_1+T_2) \right)}
\\
& = \displaystyle{(T_1+T_2)\sum\limits_{n=0}^\infty \mathbf{P}_{x_0}\left(\tau(x_0, \partial U)> n(T_1+T_2)\right)}
\\
& \displaystyle{\leq (T_1+T_2)\sum\limits_{n=0}^\infty \left[1- \min\limits_{z\in U\cup \partial U}\mathbf{P}_{z}\left(\tau(x_0, \partial U)\leq T_1+T_2\right)\right]^n}
\\
& \displaystyle{= (T_1+T_2)\left(\min\limits_{z\in U\cup \partial U}\mathbf{P}_{z}\left(\tau(x_0, \partial U)\leq T_1+T_2\right)\right)^{-1}}
\\
& \leq 2(T_1+T_2)\exp\left(\dfrac{1}{\varepsilon}(V_0+d)\right) \ ,
\end{array}
\end{equation}
as $\varepsilon>0$ is small, which establishes the upper bound of \eqref{Thm:ExponentialAsymptoticMeanExitTimeExitPosition:Eq:ExponentialExitTime}.

Now we let $\nu$ be the smallest $n$ for which $Z_n=x(\tau_n)\in \partial D$. Then by
\eqref{Thm:ExponentialAsymptoticMeanExitTimeExitPosition:Eq:ExitProbabilityUpperBound} we obtain that

\begin{equation}\label{Thm:ExponentialAsymptoticMeanExitTimeExitPosition:Eq:ExponentialExitTimeLowerBoundMarkovChain}
\mathbf{P}_{x}(\nu\geq n)\geq \left[1-\exp\left(-\dfrac{1}{\varepsilon}(V_0-d)\right)\right]^{n-1}
\end{equation}
for $x\in \gamma$. As we have $\tau(x, \partial U)=(\tau_1-\tau_0)+(\tau_2-\tau_1)+...+(\tau_{\nu}-\tau_{\nu-1})$, we obtain from
the strong Markov property of $x(t)$ that we have

\begin{equation}\label{Thm:ExponentialAsymptoticMeanExitTimeExitPosition:Eq:ExponentialExitTimeLowerBoundCycle}
\begin{array}{ll}
\mathbf{E} \tau(x, \partial U) & = \displaystyle{\sum\limits_{n=1}^\infty \mathbf{E}\left((\tau_n-\tau_{n-1})\1_{\{\nu\geq n\}}\right)}
\\
& \geq \displaystyle{\sum\limits_{n=1}^\infty \mathbf{E}\left((\tau_n-\sigma_{n-1})\1_{\{\nu\geq n\}}\right)}
\\
& \geq \displaystyle{\sum\limits_{n=1}^\infty \mathbf{P}_x(\nu\geq n) \cdot \inf\limits_{x\in \Gamma}\mathbf{E}_x \tau_1} \ .
\end{array}
\end{equation}

Since there exists some $t_1>0$ such that $\inf\limits_{x\in \Gamma}\mathbf{E}_x \tau_1\geq t_1>0$, we obtain from
\eqref{Thm:ExponentialAsymptoticMeanExitTimeExitPosition:Eq:ExponentialExitTimeLowerBoundMarkovChain} and
\eqref{Thm:ExponentialAsymptoticMeanExitTimeExitPosition:Eq:ExponentialExitTimeLowerBoundCycle} that
for $x\in \gamma$ we have
$$\mathbf{E} \tau(x, \partial U) \geq t_1\sum\limits_{n=1}^\infty \left[1-\exp\left(-\dfrac{1}{\varepsilon}(V_0-d)\right)\right]^{n-1}
\geq t_1\exp\left(\dfrac{1}{\varepsilon}(V_0-d)\right) \ .$$
For an arbitrary $x_0\in U$ we obtain that

\begin{equation}\label{Thm:ExponentialAsymptoticMeanExitTimeExitPosition:Eq:ExponentialExitTimeLowerBound}
\begin{array}{ll}
\mathbf{E} \tau(x_0, \partial U) & \geq \mathbf{E} \left[\1_{\{\tau(x(\tau_1), \partial U)>\tau_1\}}\mathbf{E}_{x(\tau_1)}(\tau(x(\tau_1), \partial U))\right]
\\
& \geq t_1\exp\left(\dfrac{1}{\varepsilon}(V_0-d)\right)\mathbf{P}_{x_0}\left(\tau(x(\tau_1), \partial U)>\tau_1\right)
\\
& \geq \dfrac{t_1}{2}\exp\left(\dfrac{1}{\varepsilon}(V_0-d)\right) \ ,
\end{array}
\end{equation}
as we have $\mathbf{P}_{x_0}\left(\tau(x(\tau_1), \partial U)>\tau_1\right)\rightarrow 1$ when $\varepsilon\rightarrow 0$. This establishes the lower bound
of \eqref{Thm:ExponentialAsymptoticMeanExitTimeExitPosition:Eq:ExponentialExitTime}. $\square$

\section{Derivation of the Hamilton--Jacobi equation for the quasi--potential.}
\label{Appendix:HJB-Quasi-Potential}

The local quasi--potential function $\phi_{\text{loc}}^{\text{QP}}(x;x_0)$ defined in \eqref{Eq:LocalQuasiPotentialVariationalProblem}
is expressed through a variational infimum over trajectories $\psi$ connecting $x_0$ to $x$:

\begin{equation}\label{Appendix:HJB-Quasi-Potential:Eq:LocalQuasiPotentialVariationalProblem}
\phi_{\text{loc}}^{\text{QP}}(x;x_0)= \inf\limits_{T>0}\inf\limits_{\psi(0)=x_0, \psi(T)=x}S_{0T}(\psi) \ .
\end{equation}

In our framework, we are considering the SGD process $x(t)$ with an explicitly constructed action functional (rate functional)
$S_{0T}(\psi)$ that takes the form \eqref{Eq:ActionFunctional}:

\begin{equation}\label{Appendix:HJB-Quasi-Potential:Eq:ActionFunctional}
S_{0T}(\psi)=\left\{\begin{array}{l}
\displaystyle{\dfrac{1}{2}\int_0^T (\dot{\psi}_t+\nabla f(\psi_t))^T D^{-1}(\psi_t)(\dot{\psi}_t+\nabla f(\psi_t)) dt} \ ,
\\
\qquad \qquad \qquad \text{if } \psi_t \text{ is differentiable for almost every}  \ t\in [0,T] \ ;
\\
+\infty \ ,
\\
\qquad \qquad \qquad \text{otherwise \ .}
\end{array}\right.
\end{equation}

The variational problem \eqref{Appendix:HJB-Quasi-Potential:Eq:LocalQuasiPotentialVariationalProblem} with the functional
$S_{0T}(\psi)$ given in \eqref{Appendix:HJB-Quasi-Potential:Eq:ActionFunctional} can be solved explicitly via standard analysis
in the calculus of variations. This leads to the Hamilton--Jacobi equation \eqref{Thm:Hamilton-JacobiEqLocalQuasiPotential:HJE}
satisfied by $\phi_{\text{loc}}^\text{QP}(x;x_0)$. Below we provide a proof of Theorem \ref{Thm:Hamilton-JacobiEqLocalQuasiPotential}
in Section \ref{Sec:LocalQuasiPotential} of the paper.

\

\textbf{Proof of Theorem \ref{Thm:Hamilton-JacobiEqLocalQuasiPotential}.}

Let $A(x)=D^{-1}(x)$. We introduce the $A$--norm $\|u\|^2_{A(x)}=u^TA(x)u$
for $u\in \mathbb{R}^d$, $x\in \mathbb{R}^d$ and $A(x)\in \mathbb{R}^d\otimes \mathbb{R}^d$, and the $A$--inner product
$\langle u, v\rangle_{A(x)}=u^TA(x)v$ for $u\in \mathbb{R}^d$, $v\in \mathbb{R}^d$, $x\in \mathbb{R}^d$ and $A(x)\in \mathbb{R}^d\otimes \mathbb{R}^d$.
For a function $\psi: [0,T] \rightarrow \mathbb{R}^d$ that is  differentiable for almost every $t\in [0, T]$, we have

$$\begin{array}{ll}
&S_{0T}(\psi)
\\
=&
\displaystyle{\dfrac{1}{2}\int_0^T \left(\dot{\psi}_t+\nabla f(\psi_t)\right)^T A(\psi_t)\left(\dot{\psi}_t+\nabla f(\psi_t)\right) dt}
\\
=&\displaystyle{\dfrac{1}{2}\int_0^T \left(\|\dot{\psi}_t\|^2_{A(\psi_t)}+ 2\langle \dot{\psi}_t, \nabla f(\psi_t)) \rangle_{A(\psi_t)}+\|\nabla f(\psi_t)\|^2_{A(\psi_t)}\right) dt}
\\
\geq &\displaystyle{\dfrac{1}{2}\int_0^T
\left(2\|\dot{\psi}_t\|_{A(\psi_t)}\|\nabla f(\psi_t)\|_{A(\psi_t)}+ 2\langle \dot{\psi}_t, \nabla f(\psi_t) \rangle_{A(\psi_t)}\right) dt}
\\
= &\displaystyle{\int_0^T
\left(\|\dot{\psi}_t\|_{A(\psi_t)}\|\nabla f(\psi_t)\|_{A(\psi_t)}+ \langle \dot{\psi}_t, \nabla f(\psi_t) \rangle_{A(\psi_t)}\right) dt} \ .
\end{array}
$$
Here we have use the inequality $\|\dot{\psi}_t\|^2_{A(\psi_t)}+\|\nabla f(\psi_t)\|^2_{A(\psi_t)}\geq 2\|\dot{\psi}_t\|_{A(\psi_t)}\|\nabla f(\psi_t)\|_{A(\psi_t)}$ where the equality is taken when $\|\dot{\psi}_t\|_{A(\psi_t)}=\|\nabla f(\psi_t)\|_{A(\psi_t)}$.
We can re--parameterize the path $\psi_t$ into $\widehat{\psi}_t$ such
that $\|\dot{\widehat{\psi}}_t\|_{A(\widehat{\psi}_t)}=
\|\nabla f(\widehat{\psi}_t)\|_{A(\widehat{\psi}_t)}$ for $0\leq t \leq T$. In this way

$$S_{0T}(\psi)\geq S(\widehat{\psi})=
\int_0^{T_{\widehat{\psi}}}
\left(\|\dot{\widehat{\psi}}_t\|_{A(\widehat{\psi}_t)}\|\nabla f(\widehat{\psi}_t)\|_{A(\widehat{\psi}_t)}
+ \langle \dot{\widehat{\psi}}_t, \nabla f(\widehat{\psi}_t) \rangle_{A(\widehat{\psi}_t)}\right) dt$$
for some $T_{\widehat{\psi}}$ that depends on $\widehat{\psi}$.

The above integral is indeed independent of the parametrization of $\widehat{\psi}_t$. In particular, we can choose the
arc--length parametrization to obtain the \textit{geometric action}

$$S(\psi)=
\int_0^L
\left(\left\|\dfrac{d\psi_s}{ds}\right\|_{A(\psi_s)}\|\nabla f(\psi_s)\|_{A(\psi_s)}
+ \left\langle \dfrac{d\psi_s}{ds}, \nabla f(\psi_s) \right\rangle_{A(\psi_s)}\right) ds \ ,$$
where $L$ is the total length of the path $\psi$ and $s$ is the arc--length parameter.
Thus \eqref{Appendix:HJB-Quasi-Potential:Eq:LocalQuasiPotentialVariationalProblem} becomes

$$
\phi_{\text{loc}}^{\text{QP}}(x;x_0)= \inf\limits_{\psi(0)=x_0, \psi(L)=x}S(\psi) \ .
$$

From the above display, we can take a parametrization such that $\|\dot{\psi}\|_{A(\psi)}\equiv 1$. We pick $\varepsilon>0$. Then
we can apply the Bellman's optimality principle to the geometric action and we have

$$\phi_{\text{loc}}^{\text{QP}}(x;x_0)=
\inf\limits_{\|\dot{\psi}\|_{A(\psi)}=1}\left\{\int_0^\varepsilon
\left(\|\nabla f(\psi)\|_{A(\psi)}+\nabla f(\psi)^TA(\psi)\dot{\psi}\right)ds+
\phi_{\text{loc}}^{\text{QP}}\left(x-\int_0^\varepsilon \dot{\psi}ds ; x_0\right)\right\} \ .$$

Apply a Taylor expansion to $\phi_{\text{loc}}^{\text{QP}}(x;x_0)$ in $x$
to $\displaystyle{\phi_{\text{loc}}^{\text{QP}}\left(x-\int_0^\varepsilon \dot{\psi}ds ; x_0\right)}$
we obtain

$$\phi_{\text{loc}}^{\text{QP}}(x;x_0)=
\inf\limits_{\|\dot{\psi}\|_{A(\psi)}=1}\left\{\varepsilon
\left(\|\nabla f(\psi)\|_{A(\psi)}+\nabla f(\psi)^TA(\psi)\dot{\psi}-
\nabla \phi_{\text{loc}}^{\text{QP}}(x; x_0)^T \dot{\psi}\right)+
\phi_{\text{loc}}^{\text{QP}}\left(x ; x_0\right)+o(\varepsilon^2)\right\} \ .$$

Cancelling $\phi_{\text{loc}}^{\text{QP}}(x;x_0)$ on both sides of the above display and divide by $\varepsilon$, then send $\varepsilon\rightarrow 0$,
we obtain

$$0=
\inf\limits_{\|\dot{\psi}\|_{A(\psi)}=1}\left\{
\|\nabla f(\psi)\|_{A(\psi)}+\nabla f(\psi)^TA(\psi)\dot{\psi}-
\nabla \phi_{\text{loc}}^{\text{QP}}(x; x_0)^T \dot{\psi}\right\} \ .$$

Notice that the term
$$\begin{array}{ll}
\nabla f(\psi)^TA(\psi)\dot{\psi}-
\nabla \phi_{\text{loc}}^{\text{QP}}(x; x_0)^T \dot{\psi}
&= -\left(-\nabla f(\psi)+
A^{-1}(\psi)\nabla \phi_{\text{loc}}^{\text{QP}}(x; x_0)\right)^TA(\psi)\dot{\psi}
\\
&= -\left\langle -\nabla f(\psi)+
A^{-1}(\psi)\nabla \phi_{\text{loc}}^{\text{QP}}(x; x_0), \dot{\psi}\right\rangle_{A(\psi)} \ .
\end{array}$$
The above term is minimal when the inner product is maximal, that is
$$\dot{\psi}=\dfrac{-\nabla f(\psi)+
A^{-1}(\psi)\nabla \phi_{\text{loc}}^{\text{QP}}(x; x_0)}{\left\| -\nabla f(\psi)+
A^{-1}(\psi)\nabla \phi_{\text{loc}}^{\text{QP}}(x; x_0)\right\|_{A(\psi)}} \ .$$
Thus

$$0=
\|\nabla f(\psi)\|_{A(\psi)}-\left\| -\nabla f(\psi)+
A^{-1}(\psi)\nabla \phi_{\text{loc}}^{\text{QP}}(x; x_0)\right\|_{A(\psi)} \ ,$$
i.e.

$$
\|\nabla f(\psi)\|_{A(\psi)}=\left\| -\nabla f(\psi)+
A^{-1}(\psi)\nabla \phi_{\text{loc}}^{\text{QP}}(x; x_0)\right\|_{A(\psi)} \ .$$
Square the above display we get

$$\nabla f(\psi)^TA(\psi)\nabla f(\psi)=
\nabla f(\psi)^TA(\psi)\nabla f(\psi)+
\left(\nabla \phi_{\text{loc}}^{\text{QP}}(x; x_0)\right)^TA^{-1}(\psi)
\nabla \phi_{\text{loc}}^{\text{QP}}(x; x_0)-2\nabla f(x)\cdot \nabla \phi_{\text{loc}}^{\text{QP}}(x; x_0) \ ,$$
where $A^{-1}(x)=D(x)$. Cancelling the $\nabla f(\psi)^TA(\psi)\nabla f(\psi)$ term
on both sides leads to \eqref{Thm:Hamilton-JacobiEqLocalQuasiPotential:HJE}. $\square$

\section{Closeness of $x(t)$ to $x^{\text{GD}}(t)$ for small $\varepsilon>0$.}
\label{Appendix:PerturbationGD-SGD}

\textbf{Proof of Lemma \ref{Lemma:SGD-GD-Closeness}.}

We can write the processes $x(t), x(0)=x_0\in \mathbb{R}^d$ and $x^{\text{GD}}(t), x^{\text{GD}}(0)=x_0\in \mathbb{R}^d$
in integral form so that we have

$$x(t)=x_0-\int_0^t \nabla f(x(s))ds+\sqrt{\varepsilon}\int_0^t \Sigma(x(s))dW(s) \ ,$$

$$x^{\text{GD}}(t)=x_0-\int_0^t \nabla f(x^{\text{GD}}(s))ds \ .$$

This gives

$$
x(t)-x^{\text{GD}}(t)=-\int_0^t \left(\nabla f(x(s))-\nabla f(x^{\text{GD}}(s))\right)ds
+\sqrt{\varepsilon}\int_0^t \Sigma(x(s))dW(s) \ .
$$

By the inequality $(a+b)^2\leq 2(a^2+b^2)$, from the above we have

\begin{equation}\label{Lemma:SGD-GD-Closeness:Eq:SquareErrorEstimateStep1}
|x(t)-x^{\text{GD}}(t)|^2 \leq \displaystyle{2\left|\int_0^t \left(\nabla f(x(s))-\nabla f(x^{\text{GD}}(s))\right)ds\right|^2
+2\varepsilon\left|\int_0^t \Sigma(x(s))dW(s)\right|^2} \ .
\end{equation}

Since $\nabla f(x)$ is $L$--Lipschitz, we can estimate

$$\left|\int_0^t \left(\nabla f(x(s))-\nabla f(x^{\text{GD}}(s))\right)ds\right|\leq
\int_0^t \left|\nabla f(x(s))-\nabla f(x^{\text{GD}}(s))\right|ds\leq
L\int_0^t \left|x(s)-x^{\text{GD}}(s)\right|ds \ .$$

By Cauchy--Schwarz inequality, we then have

$$\left|\int_0^t \left(\nabla f(x(s))-\nabla f(x^{\text{GD}}(s))\right)ds\right|^2 \leq
L^2\left(\int_0^t \left|x(s)-x^{\text{GD}}(s)\right|ds\right)^2\leq L^2t\int_0^t \left|x(s)-x^{\text{GD}}(s)\right|^2ds \ .$$

This combined with \eqref{Lemma:SGD-GD-Closeness:Eq:SquareErrorEstimateStep1} gives us that

$$
\begin{array}{ll}
|x(t)-x^{\text{GD}}(t)|^2 & \leq \displaystyle{2L^2t\int_0^t \left|x(s)-x^{\text{GD}}(s)\right|^2ds
+2\varepsilon\left|\int_0^t \Sigma(x(s))dW(s)\right|^2} \ .
\end{array}
$$

Thus for $0\leq t \leq T$ we have

\begin{equation}\label{Lemma:SGD-GD-Closeness:Eq:SquareErrorEstimateStep2}
\begin{array}{ll}
\mathbf{E}\left|x(t)-x^{\text{GD}}(t)\right|^2 & \leq \displaystyle{2L^2T\int_0^t \mathbf{E}\left|x(s)-x^{\text{GD}}(s)\right|^2ds
+2\varepsilon\mathbf{E}\left|\int_0^t \Sigma(x(s))dW(s)\right|^2}
\\
& =\displaystyle{2L^2T\int_0^t \mathbf{E}\left|x(s)-x^{\text{GD}}(s)\right|^2ds
+2\varepsilon\int_0^t \mathbf{E}\text{Tr} D(x(s))ds}
\\
& \leq \displaystyle{2L^2T\int_0^t \mathbf{E}\left|x(s)-x^{\text{GD}}(s)\right|^2ds
+2\varepsilon MT} \ .
\end{array}
\end{equation}
Here we have used \eqref{Assumption:SGDLossfDiffusionD:Eq:DiffusionDBoundedTrace} in
the Assumption \ref{Assumption:SGDLossfDiffusionD} and the It\^{o}
isometry (see \cite[Corollary 3.1.7]{[OksendalSDE]}). The estimate \eqref{Lemma:SGD-GD-Closeness:Eq:SquareErrorEstimateStep2}
and the Gronwall inequality (see \cite[Exercise 5.17]{[OksendalSDE]}) give us that for $0\leq t \leq T$,

$$\mathbf{E}\left|x(t)-x^{\text{GD}}(t)\right|^2
 \leq 2\varepsilon MT e^{L^2Tt}$$

Thus

$$\max\limits_{0\leq t \leq T}\mathbf{E}\left|x(t)-x^{\text{GD}}(t)\right|^2
 \leq 2\varepsilon MT e^{L^2T^2}=C\varepsilon$$
where $C=C(M,T,L)=2MTe^{L^2T^2}$. This is what is stated in Lemma \ref{Lemma:SGD-GD-Closeness}. $\square$

\end{appendix}

\end{document}